\documentclass[a4paper,fleqn]{cas-sc}

\usepackage[authoryear,longnamesfirst]{natbib}
\usepackage{multirow}
\usepackage{booktabs}
\usepackage{tcolorbox}

\def\tsc#1{\csdef{#1}{\textsc{\lowercase{#1}}\xspace}}
\tsc{WGM}
\tsc{QE}
\tsc{EP}
\tsc{PMS}
\tsc{BEC}
\tsc{DE}

\begin{document}
\let\WriteBookmarks\relax
\renewcommand{\topfraction}{0.95}
\renewcommand{\bottomfraction}{0.85}
\setcounter{topnumber}{4}
\setcounter{bottomnumber}{2}
\setcounter{totalnumber}{6}
\renewcommand{\textfraction}{0.05}
\renewcommand{\floatpagefraction}{0.80}
\renewcommand{\dbltopfraction}{0.95}
\renewcommand{\dblfloatpagefraction}{0.80}

\shorttitle{Definition-Guided In-Context Learning for SLU}
\shortauthors{N. Agrawal and S. Ganapathy}

\newcommand{\method}{DG-ICL}                                  
\newcommand{\methodexp}{Definition-Guided In-Context Learning}
\newcommand{\methodexpbold}{\textbf{D}efinition-\textbf{G}uided \textbf{I}n-\textbf{C}ontext \textbf{L}earning}
\newcommand{\methodsym}{DG-ICL-Sym}                           
\newcommand{\methodflip}{DG-ICL-Flip}                         

\title[mode=title]{Definition-Guided In-Context Learning for Zero-Shot Spoken Language Understanding}                    



\author[1]{Neeraj Agrawal}
\ead{aneeraj@iisc.ac.in}

\author[1]{Sriram Ganapathy}
\cormark[1]
\ead{sriramg@iisc.ac.in}

\affiliation[1]{organization={Department of Electrical Engineering, Indian Institute of Science},
                city={Bengaluru}, 
                postcode={560012}, 
                state={Karnataka}, 
                country={India}}

\cortext[cor1]{Corresponding author}

\begin{abstract}
Adapting spoken language understanding (SLU) systems to novel or revised label taxonomies conventionally requires additional labeled data and retraining, creating a bottleneck for real-world deployment. While in-context learning with few-shot demonstrations offers an alternative, multi-modal generalization and scaling has been challenging.
In this work, we investigate whether natural-language definitions alone can guide zero-shot SLU without demonstration exemplars, finding that symbolic definitions serve as the primary supervisory driver for adapting speech-text models. We operationalize this principle as \methodexpbold{} (\method{}), a fine-tuning framework that decouples semantic categories from fixed lexical names. By replacing labels with dynamically randomized definitions (arbitrary synthetic tokens in \methodsym{} or permuted vocabulary words in \methodflip{}), \method{} avoids lexical shortcuts and encourages the model to ground acoustic representations in contextual definitions. Novel categories can thus be introduced at inference simply by defining them through the instructions. Across three speech-text models  (\textsc{Audio Flamingo~3}, \textsc{Qwen2-Audio}, and \textsc{SALMONN}), \method{} outperforms standard instruction tuning on held-out tasks under novel symbols ($0.663$ vs.\ $0.449$ macro-F1 on Audio Flamingo~3; $0.703$ vs.\ $0.620$ on Qwen2-Audio; $0.582$ vs.\ $0.399$ on SALMONN), while remaining competitive on original labels. Furthermore, counterfactual evaluations demonstrate instruction following abilities, where it is shown that \methodsym{} retains $68\%$ of its performance when definitions contradict pre-trained lexical priors, whereas standard instruction tuning elicits significantly reduced retention of $21\%$. Finally, we show that \method{} enables zero-shot speech clustering into user-defined categories specified on the fly.
\end{abstract}

\begin{keywords}
Spoken language understanding \sep
In-context learning \sep
Zero-shot transfer \sep
Speech-text foundation models \sep
Instruction following
\end{keywords}

\maketitle

\section{Introduction}
\label{sec:intro}
Spoken language understanding (SLU) enables conversational systems to move beyond speech transcription toward decoding the underlying meaning and intent in speech. SLU encompasses diverse tasks, ranging from user intent classification~\cite{minds14} and dialogue-act parsing~\cite{wu2021harpervalleybankdomainspecificspokendialog} to speech emotion recognition~\cite{cremad} and keyword spotting~\cite{speechcommands}. Resolving these tasks requires interpreting acoustic and prosodic nuances directly from continuous speech rather than from  intermediate transcripts. However, progress in SLU remains constrained by the cost of acquiring task-specific audio annotations. This burden compounds in practical deployments, where expanding to new domains or updating category taxonomies necessitates a cycle of data re-annotation and model retraining, rendering conventional supervised pipelines ineffective on unseen tasks.

\begin{figure*}
  \centering
  \includegraphics[width=\linewidth]{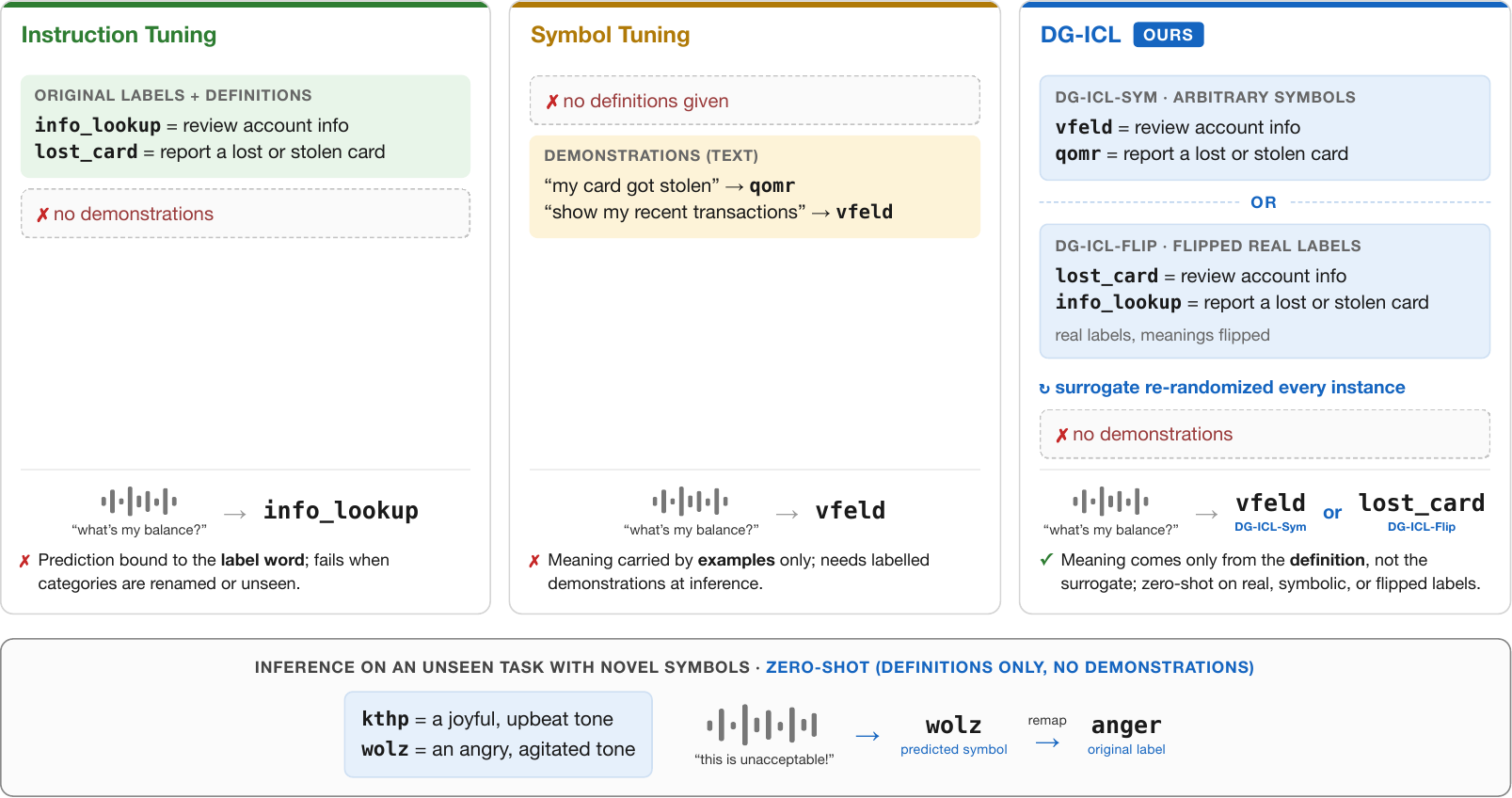}
 \caption{The three fine-tuning paradigms share the same spoken input and the same frozen speech-text LLM (adapted with LoRA); they differ only in how a category's defintion is conveyed. \textbf{Instruction Tuning} keeps the original label, so predictions bind to that label and fail when categories are renamed or unseen. \textbf{Symbol Tuning}~\cite{wei2023symbol} replaces labels with arbitrary symbols whose meaning is supplied only through few-shot demonstrations, with no definitions. \textbf{\method{}} (ours) instead conveys each category's label through a natural-language \emph{definition} and it evaluates without demonstrations,
randomizing the label-surrogate assignment on every training instance so the symbol never acquires a fixed meaning; it comes in two forms, \methodsym{} (arbitrary symbols) and \methodflip{} (deranged real labels, whose usual meaning is reassigned by the definition). At inference (bottom), a previously unseen task with novel symbols is introduced zero-shot by describing its categories:
the model predicts a symbol, which is then remapped to the underlying category (for evaluation scoring against ground-truth labels).}
  \label{fig:overview}
\end{figure*}

Historically, SLU relied on cascaded pipelines coupling an Automatic Speech Recognition (ASR) front-end with a downstream text-based NLU module~\cite{peng2023study}. These approaches can propagate transcription errors and discard expressive prosodic cues~\cite{serdyuk2018towards}. While end-to-end architectures directly map waveforms to semantic targets~\cite{saxon2021end}, they remain constrained by static classification heads that cannot adapt to altered label spaces without gradient updates. To overcome these constraints, recent advances converge on speech-text large language models (speech-text LLMs) that jointly ingest audio representations and textual tokens within a unified autoregressive model~\cite{chu2024qwen2, tang2024salmonn, goel2025audioflamingo3advancing}.

In practical applications, speech-text LLMs are often required to classify audio into custom categories defined on the fly, where collecting large volume labeled audio demonstrations for each class may be impractical. While foundation models exhibit flexible in-context learning (ICL) in text~\cite{dong2024survey, wei2021finetuned, lampinen-etal-2022-language}, extending this paradigm to multimodal speech-text settings remains fragile~\cite{alayrac2022flamingo, doveh2024towards, wang2024can}.   While conventional instruction tuning improves accuracy on familiar tasks, it introduces a shortcut: the model learns to memorize the lexical names of familiar labels rather than grounding speech in the accompanying definitions~\cite{wei2023symbol}. Consequently, performance degrades sharply whenever category taxonomies are novel, renamed, or abstracted into symbols. Prior methods attempt to decouple labels from class definitions via symbol tuning~\cite{wei2023symbol} or label flipping~\cite{agrawal2025spokenlanguageunderstandingunseen}, but both approaches require few-shot demonstration exemplars.

The key problem statement of this work - \emph{are demonstrations necessary to convey a label's meaning, or does a concise natural-language definition suffice on its own?} We address it with \methodexpbold{} (\method{}), a fine-tuning framework for speech-text LLMs (Figure~\ref{fig:overview}) in which   labels are grounded \emph{solely} through natural-language definitions, without any demonstrations (few-shot examples) at inference ($k=0$). Crucially, the label-to-symbol mapping is dynamically re-randomized for every training instance. This continuous variation eliminates static lexical shortcuts and encourages the model to resolve speech categories through the accompanying instructions. We instantiate this decoupling in two complementary forms: replacing labels with arbitrary synthetic symbols (\methodsym{}) for abstracted taxonomies, and permuting semantic labels (\methodflip{}) to preserve original vocabulary alignment. Following the principle of learning to execute tasks from context~\cite{hsu2024meta}, we train \method{} across diverse SLU tasks (dialogue acts, emotion, intent) and evaluate   on held-out tasks and unseen acoustic domains (regional accents, sung prosody, and cross-lingual transfer). In doing so, \method{} extends our   single-task study~\cite{agrawal2025spokenlanguageunderstandingunseen} into a unified, demonstration-free, multi-task framework capable of categorizing speech into user-specified schemes on demand.

The primary contributions of this work are as follows:
\begin{itemize}
    \item \textbf{Definitions in place of demonstrations.} We show that natural-language category definitions match or surpass few-shot demonstrations for speech classification, removing the need for labeled speech exemplars at inference ($k=0$).

    \item \textbf{\methodexp{} (\method).} We introduce a fine-tuning framework that decouples label tokens from fixed semantics by binding categories to random symbols (\methodsym{}) or permuted names (\methodflip{}).

    \item \textbf{Counterfactual steerability and definition tracking.} We find that \method{}-tuned models ground their predictions in context, following definitions even under semantic contradiction.

    \item \textbf{Zero-shot user-defined clustering.} We demonstrate that \method{} enables a single model to sort speech into novel, user-specified taxonomy categories for both spoken utterances and written text queries, remaining robust under symbolic labels.

    \item \textbf{Generalization and consistency.} We validate consistent zero-shot gains across three distinct speech-text models (\textsc{Audio Flamingo 3}, \textsc{Qwen2-Audio}, and \textsc{SALMONN}), and verify that this specialization retains generative spoken question answering and keyword spotting capabilities.
\end{itemize}
To facilitate reproducibility, our complete code-base, prompts, and training configurations are made publicly available.\footnote{\url{https://github.com/MagnIeeT/ICL-speech-text-LLM}}

\section{Related Work}
\label{sec:related-work}

\noindent\textbf{Few-Shot Learning and In-Context Learning}: In-context learning (ICL) allows autoregressive language models to execute unseen tasks at inference time via prompt demonstration exemplars~\cite{DBLP, gargshivam}. While this capability has been extensively studied in text~\cite{razumovskaia2024analyzing, shiwei, liu2022few, gargshivam, 1049803}, extending in-context learning to multimodal speech-text settings remains challenging. In speech, models exhibit a strong bias toward textual representations while under-utilizing continuous acoustic cues~\cite{wang2024can}. Prior efforts attempt to elicit speech ICL by adapting pre-trained audio encoders~\cite{wang2024can}, exploring acoustic demonstrations in textless settings~\cite{chang2024exploring}, or by applying multimodal instruction tuning~\cite{pan2023cosmic, li2023instruction}. However, they remain reliant on curated, context-heavy demonstrations to convey category mappings. In contrast, our work establishes that novel tasks with new class definitions can be introduced  \emph{without demonstration exemplars} ($k=0$), relying exclusively on concise natural-language definitions.

\noindent\textbf{Cross-Task Generalization in Zero-Shot and Multi-Task SLU}: Multi-task training enhances representational robustness by encouraging models to learn shared, transferable abstractions~\cite{zhang2021survey}, while cross-task in-context evaluation assesses out-of-distribution adaptation~\cite{chatterjee2024language}. In speech processing, evaluating on held-out tasks across diverse acoustic domains provides an established benchmark for cross-domain generalization~\cite{shon2022slue, huang2023zero}. Complementary strategies such as dynamic prompt retrieval~\cite{cheng-etal-2023-uprise}, meta-learning (e.g., MetaICL~\cite{min2021metaicl} and context-execution meta-training~\cite{hsu2024meta}), show that conditioning on diverse tasks fosters context understanding. We adopt a multi-task, held-out evaluation protocol across distinct task families, showing that direct multi-task fine-tuning paired with \method{} establishes robust, zero-shot capabilities.

\noindent\textbf{Symbolic and Abstract Representations for ICL}:  In text, instruction tuning encourages models to attend to task descriptions~\cite{li2023instruction}, explanations improve exemplar following~\cite{lampinen-etal-2022-language}, and symbol tuning formalizes label decoupling by substituting arbitrary tokens for semantic classes~\cite{wei2023symbol}. Crucially, symbol tuning conveys symbol meanings solely through few-shot demonstrations. This dependence on exemplars persists in multimodal extensions: SymDPO~\cite{jia2025symdpo} grounds symbolic targets via visual exemplars, while our prior speech work~\cite{agrawal2025spokenlanguageunderstandingunseen} required   transcript demonstrations alongside permuted labels. Consequently, existing approaches are dependent on labeled demonstration exemplars.

\noindent\textbf{Positioning This Work}: Unlike prior symbol tuning~\cite{wei2023symbol, jia2025symdpo} or single-task label flipping~\cite{agrawal2025spokenlanguageunderstandingunseen} that depend on demonstration exemplars or ASR retrieval, \method{} grounds category semantics \emph{strictly through natural-language definitions}. By dynamically pairing definitions with synthetic symbols (\methodsym{}) or permuted vocabulary labels (\methodflip{}), \method{} achieves zero-shot capabilities  across diverse speech-text LLMs.

\section{Methodology}
\label{sec:methodology}

We first define our problem formulation and reference baselines (Sections~\ref{subsec:formulation}--\ref{subsec:baselines}), and then formalize the \method{} training and inference framework (Section~\ref{subsec:framework}).

\subsection{Problem Formulation and Notation}
\label{subsec:formulation}

Let $\mathcal{T}$ denote a universe of SLU tasks, partitioned into disjoint subsets of training tasks $\mathcal{T}_{\text{FT}} \subset \mathcal{T}$ and held-out evaluation tasks $\mathcal{T}_{\text{E}} \subset \mathcal{T}$, such that $\mathcal{T}_{\text{FT}} \cap \mathcal{T}_{\text{E}} = \emptyset$. Each task $T \in \mathcal{T}$ is associated with a dataset $\mathcal{D}_T = \{(x^{(j)}, y^{(j)})\}_{j=1}^{|\mathcal{D}_T|}$, where $x \in \mathcal{X}$ denotes an input instance (either a raw speech utterance for audio queries, or a textual input) and $y \in L_{\text{sem}}^T$ represents the ground-truth class from the discrete semantic label inventory $L_{\text{sem}}^T = \{l_1, \dots, l_{|L_{\text{sem}}^T|}\}$.

For any task $T$, the task definition  $I(T)$ is defined as:
\begin{equation}
\label{eq:legend}
I(T) = \Big( \mathrm{instr}(T),\, \big\{ (l,\, \mathrm{def}(l)) \big\}_{l \in L_{\text{sem}}^T} \Big),
\end{equation}
which unifies an overall task instruction $\mathrm{instr}(T)$ with an explicit natural-language definition $\mathrm{def}(l)$ for each canonical label $l \in L_{\text{sem}}^T$. The input prompt $P(x)$ presented to the model is formatted as:
\begin{equation}
\label{eq:prompt_sem}
P(x) = \big[\, I(T) \;;\; \mathcal{E}_k \;;\; x \,\big],
\end{equation}
where $\mathcal{E}_k = \big( (x_1, y_1), \dots, (x_k, y_k) \big)$ denotes an optional sequence of $k$ textual demonstration (in-context examples) pairs (where each input $x_i$ is a text transcript), and $x$ denotes the target evaluation query (a raw speech utterance by default). In the zero-shot regime ($k=0$),  $P(x) = [\, I(T) \;;\; x \,]$. Given prompt $P(x)$, the speech-text LLM generates the target label.

\subsection{Reference Fine-Tuning Strategies}
\label{subsec:baselines}

To isolate the precise origins of definition-guided generalization gains, we compare \method{} against four reference configurations:
\begin{itemize}
    \item \textbf{Pre-Trained Foundation Models:} The open-weight models, which are publicly available, are evaluated zero-shot with semantic definitions via Eq.~\eqref{eq:prompt_sem}, serving as the baseline prior to task adaptation.
    \item \textbf{Instruction Tuning (IT):} Standard supervised multi-task fine-tuning directly on canonical semantic labels $y \in L_{\text{sem}}^T$ using definition-guided zero-shot prompts from Eq.~\eqref{eq:prompt_sem}.
    \item \textbf{Symbol Tuning}~\cite{wei2023symbol}: Canonical semantic labels are replaced by arbitrary synthetic symbols $L_{\text{sym}}$,   across prompts. Category meanings are conveyed strictly through $k \ge 1$ demonstrations per class without textual definitions.
    \item \textbf{Flipped-Label Tuning}~\cite{agrawal2025spokenlanguageunderstandingunseen}: Real semantic labels are permuted with meanings supplied by definitions, reinforced with variable numbers of few-shot demonstrations ($0 \le m \le 5$)).
\end{itemize}

\subsection{The \method{} Framework}
\label{subsec:framework}

Training instances are sampled from a multi-task mixture across diverse training tasks $T \in \mathcal{T}_{\text{FT}}$ spanning disparate domains, label hierarchies, and  objectives (e.g., intent classification, dialogue acts, and emotion recognition). Multi-task gradients incentivize the shared adapter to acquire task-agnostic instruction grounding rather than task-specific decision boundaries~\cite{wei2021finetuned}.

\paragraph{Surrogate Label Tuning.}
In \method{}, each target class is dynamically assigned a label drawn from a candidate set $L_{\text{surr}}$ and specified exclusively through its in-context definition. We explore two variants:
\begin{enumerate}
    \item \textbf{\methodsym{}:} $L_{\text{surr}} = L_{\text{sym}}$ consists of synthetic alphanumeric strings with two sub-word tokens to preserve uniform target sequence lengths across classes. We pre-compute a pool of $K=100$ distinct bijective mappings $\mathcal{F}_{\text{sym}} = \{f_1, f_2, \dots, f_K\}$, where each mapping $f_k : L_{\text{sem}}^T \to L_{\text{sym}}$ assigns a distinct synthetic symbol to each ground-truth class, and sample $f \sim \mathcal{U}(\mathcal{F}_{\text{sym}})$ uniformly per training instance.
    \item \textbf{\methodflip{}:} $L_{\text{surr}} = L_{\text{sem}}^T$, where label assignments are generated on the fly for each instance via a random permutation $f \in \operatorname{Perm}(L_{\text{sem}}^T)$. This sets label semantics into frequent conflict with definition semantics across instances.
\end{enumerate}

\paragraph{Prompt Construction and Training.}
Given active mapping $f$, the surrogate definition  is formulated as:
\begin{equation}
\label{eq:surrogate_legend}
I_f(T) = \Big( \mathrm{instr}(T),\, \big\{ (f(l),\, \mathrm{def}(l)) \big\}_{l \in L_{\text{sem}}^T} \Big),
\end{equation}
which combines each label token $f(l)$ to the definition of its category. The training prompt $P_f(x)$ is constructed zero-shot ($k=0$) as:
\begin{equation}
\label{eq:prompt_sym}
P_f(x) = \big[\, I_f(T) \;;\; x \,\big].
\end{equation}
The adapter parameters  are trained using standard autoregressive cross-entropy loss to predict the surrogate tokens $f(y)$ given prompt $P_f(x)$ across all training tasks $\mathcal{T}_{\text{FT}}$.

\paragraph{Zero-Shot Inference.}
During inference on an unseen evaluation task $T \in \mathcal{T}_{\text{E}}$, the model is presented with a target label set $L^T$. The evaluation prompt $P_{\text{eval}}(x)$ is constructed using the task instruction and definition   according to Eq.~\eqref{eq:prompt_sem} (for semantic labels) or Eq.~\eqref{eq:prompt_sym} (for surrogate labels). Predictions $\hat{y}$ are obtained via greedy autoregressive decoding. When evaluating under surrogate label schemes, the generated string is mapped back to the semantic category via inverse mapping $f^{-1}(\hat{y})$ (outputs outside the active inventory are treated as invalid predictions, as detailed in Section~\ref{subsec:eval}).

\section{Experimental Setup}
\label{sec:setup}

\subsection{Datasets and Tasks}
\label{subsec:datasets}
All audio recordings are standardized and resampled to 16\,kHz. Dataset split statistics, audio duration distributions, and label definitions are detailed in Appendix~\ref{app:datasets}. To evaluate cross-task, cross-accent, and cross-lingual generalization, we consider an SLU benchmark suite across diverse task paradigms:
\begin{itemize}
    \item \textbf{Fine-tuning tasks:} Dialogue-act parsing (HVB~\cite{wu2021harpervalleybankdomainspecificspokendialog}), speech emotion recognition (CREMA-D~\cite{cremad}), and English banking intent (MInDS-14~en~\cite{minds14}).
    \item \textbf{Held-out tasks:} Indian-accented banking intent (Skit-S2I~\cite{skit}), sung-emotion recognition (RAVDESS-Song~\cite{ravdess}), and cross-lingual intent classification (MInDS-14 French and Korean)\footnote{SLUE VoxCeleb and VoxPopuli are excluded as both appear in foundation model pre-training corpora.}.
    \item \textbf{Zero-shot speech clustering:} Fine-grained intents from Skit-S2I and MInDS-14 (fr/ko) are grouped into three coarse categories via natural-language definitions across three naming styles: \emph{Meaningful} (domain-descriptive), \emph{Neutral} (generic placeholders), and \emph{Symbolic} (arbitrary character sequences). The full category names, definitions, and mapping dictionaries are reported in Appendix~\ref{app:clustering}.
    \item \textbf{Capability probes:} Out-of-domain keyword spotting without definitions (Google Speech Commands~\cite{speechcommands}) and generative spoken question answering (HeySQuAD~\cite{heysquad}).
\end{itemize}

\subsection{Models and Training Setup}
\label{subsec:models}

We apply \method{} across three   speech-text LLMs that span diverse audio fusion mechanisms:
\begin{itemize}
    \item \textbf{\textsc{Audio Flamingo~3} (AF3)}~\cite{goel2025audioflamingo3advancing} ($\sim$8B parameters)
    \item \textbf{\textsc{Qwen2-Audio}}~\cite{chu2024qwen2} ($\sim$8B parameters)
    \item \textbf{\textsc{SALMONN}}~\cite{tang2024salmonn} ($\sim$13B parameters)
\end{itemize}
To preserve foundational multimodal representations while enabling parameter-efficient tuning, pre-trained acoustic encoders and language model weights remain frozen.  The adaptation is performed via LoRA~\cite{Hu2021LoRALA} on transformer projection and attention layers.

For \method{}, each training instance is formatted with dynamic label-surrogate mappings following Section~\ref{subsec:framework}. Models are optimized using AdamW with a cosine learning rate schedule. Checkpoints are selected via mean zero-shot macro-F1 across validation splits of the fine-tuning tasks using original semantic labels. The full set of hyper-parameters (learning rates, batch sizes, LoRA rank $r$, and hardware configurations) are detailed in Appendix~\ref{app:hyperparams}.

\subsection{Evaluation Protocol and Metrics}
\label{subsec:eval}

\paragraph{Evaluation Metrics.} For classification tasks, we report macro-averaged F1 (macro-F1), where generated strings are matched against candidate target labels, and invalid generations are penalized as misclassifications (contributing zero true positives). For generative spoken QA (\textsc{HeySQuAD}), we report Exact Match (EM) and token-level F1, while for keyword spotting (Google Speech Commands), we report classification accuracy. Models are evaluated across canonical semantic names, novel abstract symbols (to test prior-free definition following), and permuted counterfactual labels (Section~\ref{subsec:steer}).

\paragraph{Statistical Rigor.} Evaluations on our primary model (\textsc{Audio Flamingo~3} and \textsc{Qwen2-Audio}) are reported as mean $\pm$ standard deviation across three random seeds with deterministic greedy decoding ($Temp{=}0$). Statistical significance is evaluated via two-sample Welch's $t$-tests with small-sample Hedges'  effect size correction ($N=3$). Due to compute constraints, the 13B \textsc{SALMONN} model is fine-tuned on a data subset from a single seed as a scale reference. The per-task training details are reported  in Appendix~\ref{app:full}.

\section{Results and Discussion}
\label{sec:results}

We empirically investigate zero-shot held-out generalization (Section~\ref{subsec:main}), the relative contribution of definitions versus demonstrations (Section~\ref{subsec:defvsdemo}), counterfactual robustness against semantic priors (Section~\ref{subsec:steer}), speech clustering into dynamic user-defined categories (Section~\ref{subsec:cluster}), and ablation analyses isolating the underlying mechanisms (Section~\ref{subsec:ablation}). Unless noted otherwise, all results reflect test-split macro-averaged F1 (macro-F1).

\subsection{Zero-Shot Generalization on Held-Out Tasks}
\label{subsec:main}

\begin{table}[t]
\centering
\small
\setlength{\tabcolsep}{4pt}
\caption{Held-out zero-shot performance (macro-F1) under definition-based evaluation.   Best symbolic average per backbone in \textbf{bold}. AF3 and Qwen2-Audio report mean\,{\scriptsize($\pm$std)} over three seeds; SALMONN is evaluated from a single seed as a scale reference. $^\dagger$ ($p < 0.01$ over strongest baseline) and $^*$ ($p < 0.01$ over Instruction Tuning) denote statistical significance via Welch's $t$-test.}
\label{tab:main}
\begin{tabular}{lcccccc}
\toprule
& \multicolumn{4}{c}{Symbolic labels} & \multicolumn{2}{c}{Avg.} \\
\cmidrule(lr){2-5}\cmidrule(lr){6-7}
Method & Skit-S2I & RAVDESS & MInDS-fr & MInDS-ko & Symbolic & Original \\
\midrule
\multicolumn{7}{l}{\textit{Audio Flamingo 3}} \\
Pre-trained          & 0.462 & 0.095 & 0.528 & 0.193 & 0.320 & 0.630 \\
Instruction Tuning & 0.503 & 0.288 & 0.549 & 0.457 & 0.449\,{\scriptsize($\pm$.041)} & 0.772\,{\scriptsize($\pm$.002)} \\
Symbol Tuning~\cite{wei2023symbol} & 0.630 & 0.124 & 0.668 & 0.417 & 0.460\,{\scriptsize($\pm$.018)} & 0.692\,{\scriptsize($\pm$.016)} \\
Flipped-Label Tuning~\cite{agrawal2025spokenlanguageunderstandingunseen} & 0.728 & 0.421 & 0.583 & 0.419 & 0.538\,{\scriptsize($\pm$.025)} & 0.776\,{\scriptsize($\pm$.005)} \\
\methodflip{} (ours)     & 0.661 & 0.470 & 0.572 & 0.434 & 0.534\,{\scriptsize($\pm$.029)} & 0.780\,{\scriptsize($\pm$.005)} \\
 \methodsym{}  (ours)   & 0.825 & 0.492 & 0.848 & 0.488 & \textbf{0.663}$^\dagger$\,{\scriptsize($\pm$.021)} & 0.766\,{\scriptsize($\pm$.026)} \\
\midrule
\multicolumn{7}{l}{\textit{Qwen2-Audio}} \\
Pre-trained          & 0.744 & 0.592 & 0.387 & 0.146 & 0.467 & 0.553 \\
Instruction Tuning & 0.810 & 0.485 & 0.743 & 0.443 & 0.620\,{\scriptsize($\pm$.017)} & 0.723\,{\scriptsize($\pm$.006)} \\
Symbol Tuning~\cite{wei2023symbol} & 0.455 & 0.319 & 0.359 & 0.262 & 0.349\,{\scriptsize($\pm$.038)} & 0.688\,{\scriptsize($\pm$.006)} \\
Flipped-Label Tuning~\cite{agrawal2025spokenlanguageunderstandingunseen} & 0.733 & 0.597 & 0.850 & 0.447 & 0.657\,{\scriptsize($\pm$.047)} & 0.657\,{\scriptsize($\pm$.007)} \\
\methodflip{} (ours)      & 0.803 & 0.645 & 0.866 & 0.454 & 0.692\,{\scriptsize($\pm$.013)} & 0.667\,{\scriptsize($\pm$.019)} \\
 \methodsym{}  (ours)   & 0.768 & 0.668 & 0.856 & 0.521 & \textbf{0.703}$^*$\,{\scriptsize($\pm$.023)} & 0.633\,{\scriptsize($\pm$.007)} \\
 \midrule

\multicolumn{7}{l}{\textit{SALMONN}} \\
Pre-trained          & 0.029 & 0.116 & 0.031 & 0.030 & 0.052 & 0.088 \\
Instruction Tuning   & 0.506 & 0.223 & 0.522 & 0.347 & 0.399 & 0.598 \\
Symbol Tuning~\cite{wei2023symbol} & 0.146 & 0.083 & 0.091 & 0.018 & 0.084 & 0.177 \\
Flipped-Label Tuning~\cite{agrawal2025spokenlanguageunderstandingunseen} & 0.560 & 0.143 & 0.589 & 0.502 & 0.448 & 0.536 \\
\methodflip{} (ours)  & 0.671 & 0.197 & 0.730 & 0.622 & 0.555 & 0.591 \\
\methodsym{} (ours)   & 0.665 & 0.295 & 0.766 & 0.604 & \textbf{0.582} & 0.563 \\
 
\bottomrule
\end{tabular}
\end{table}

Table~\ref{tab:main} presents held-out zero-shot performance across our benchmark suite under the \emph{symbolic} label mode, where models must classify unseen speech queries strictly by interpreting in-context natural-language definitions without demonstration exemplars ($k=m=0$). On our primary backbone, \textsc{Audio Flamingo~3} (AF3), \methodsym{} achieves the highest held-out average macro-F1 (\textbf{0.663}), establishing a substantial $+0.214$ margin over standard Instruction Tuning ($0.449$), while outperforming exemplar-based Symbol Tuning~\cite{wei2023symbol} ($0.460$) and the pre-trained foundation model ($0.320$). 

Three primary findings highlight the effectiveness of definition-guided decoupling:
\begin{enumerate}
    \item \textbf{Lexical shortcut reliance in standard baselines:} Although Instruction Tuning includes definitions during training, static canonical names allow the model to exploit familiar lexical shortcuts rather than grounding in definitions. Conversely, Symbol Tuning lacks definition-conditioning and cannot interpret descriptions in zero-shot settings. \methodsym{} resolves this by forming   representations   directly through definition semantics.
    
    \item \textbf{Quantifying the lexical reliance:} Comparing symbolic against original-label performance (Table~\ref{tab:main}, rightmost column) reveals this shortcut dependency. When canonical names are withheld on AF3, Instruction Tuning experiences a sharp drop of $0.323$ macro-F1 points ($0.772 \rightarrow 0.449$), retaining only $58\%$ of its performance. In contrast, \methodsym{} retains $87\%$ ($0.766 \rightarrow 0.663$), demonstrating that its decision boundaries are anchored in definition rather than lexical label choices.
    
    \item \textbf{Versatility across symbols   (\methodflip{}):} \methodflip{} attains held-out symbolic averages of $0.534$ on AF3, $0.692$ on Qwen2-Audio, and $0.555$ on \textsc{SALMONN}, consistently exceeding standard Instruction Tuning ($0.449$, $0.620$, and $0.399$) and Symbol Tuning ($0.460$, $0.349$, and $0.084$). This confirms that generalization advantages appears from the principle of definition-guided decoupling itself rather than specific synthetic token strings. Furthermore, as \methodflip{} is fine-tuned on natural-language words, it preserves strong canonical alignment ($0.780$ on AF3 vs.\ $0.772$ for Instruction Tuning).
\end{enumerate}

\paragraph{Task-Specific Generalization Patterns.}
Individual held-out tasks in Table~\ref{tab:main} reveal consistent advantages under diverse distribution shifts. On cross-lingual intent transfer, despite training exclusively on English speech, \methodsym{} attains $0.848$ on MInDS-14 French and $0.488$ on Korean (AF3), substantially outperforming Instruction Tuning ($0.549$ and $0.457$). Under acoustic shift on sung emotional speech (RAVDESS-Song), baselines degrade sharply (Instruction Tuning: $0.288$ on AF3), while \methodsym{} ($0.492$) and \methodflip{} ($0.470$) maintain strong discriminative capacity. On Indian-accented banking speech (Skit-S2I), \methodsym{} improves by $+0.322$ over Instruction Tuning ($0.825$ vs.\ $0.503$).

\paragraph{Cross-Architecture Replication on Qwen2-Audio and SALMONN.}
The advantages of definition-guided tuning replicate consistently across both models (Table~\ref{tab:main}, lower blocks). On \textsc{Qwen2-Audio}, \methodsym{} achieves the highest held-out symbolic average (\textbf{0.703}), closely followed by \methodflip{} ($0.692$), while Instruction Tuning attains $0.620$ and Symbol Tuning drops to $0.349$. The same ordering holds on the 13B-parameter \textsc{SALMONN}: \methodsym{} again attains the highest held-out symbolic average (\textbf{0.582}), followed by \methodflip{} ($0.555$), both outperforming Instruction Tuning ($0.399$), while Symbol Tuning drops to $0.084$.\footnote{Without fine-tuning, \textsc{SALMONN} tends to echo the full ``label: definition'' legend entry; these non-conforming generations are counted as invalid under our exact-match protocol, yielding a near-zero score.} These results indicate broad architectural generality offered by the proposed DG-ICL framework.

\paragraph{In-Domain Retention on Fine-Tuning Tasks.}
\begin{table}[t]
\centering
\small
\setlength{\tabcolsep}{4pt}
\caption{Trained-task performance (macro-F1) under definition-based evaluation on the three fine-tuning tasks. Best symbolic average per backbone in \textbf{bold}. Audio Flamingo~3 and Qwen2-Audio report mean\,{\scriptsize($\pm$std)} over three seeds; SALMONN is evaluated from a single seed as a scale reference. $^*$ denotes statistically significant improvement over standard Instruction Tuning ($p < 0.05$ on AF3, $p < 0.001$ on Qwen2-Audio).}
\label{tab:trained}
\begin{tabular}{lccccc}
\toprule
& \multicolumn{3}{c}{Symbolic labels} & \multicolumn{2}{c}{Avg.} \\
\cmidrule(lr){2-4}\cmidrule(lr){5-6}
Method & HVB & CREMA-D & MInDS-en & Symbolic & Original \\
\midrule
\multicolumn{6}{l}{\textit{Audio Flamingo 3}} \\
Pre-trained          & 0.083 & 0.504 & 0.415 & 0.334 & 0.517 \\
Instruction Tuning & 0.204 & 0.728 & 0.593 & 0.508\,{\scriptsize($\pm$.028)} & 0.836\,{\scriptsize($\pm$.002)} \\
Symbol Tuning~\cite{wei2023symbol} & 0.132 & 0.269 & 0.588 & 0.329\,{\scriptsize($\pm$.039)} & 0.529\,{\scriptsize($\pm$.041)} \\
Flipped-Label Tuning~\cite{agrawal2025spokenlanguageunderstandingunseen} & 0.483 & 0.497 & 0.622 & 0.534\,{\scriptsize($\pm$.048)} & 0.828\,{\scriptsize($\pm$.004)} \\
\methodflip{} (ours)     & 0.459 & 0.482 & 0.696 & 0.546\,{\scriptsize($\pm$.053)} & 0.817\,{\scriptsize($\pm$.004)} \\
\methodsym{}  (ours)   & 0.491 & 0.604 & 0.819 & \textbf{0.638}$^*$\,{\scriptsize($\pm$.044)} & 0.762\,{\scriptsize($\pm$.006)} \\
\midrule
\multicolumn{6}{l}{\textit{Qwen2-Audio}} \\
Pre-trained           & 0.030 & 0.564 & 0.511 & 0.368 & 0.452 \\
Instruction Tuning & 0.313 & 0.718 & 0.877 & 0.636\,{\scriptsize($\pm$.002)} & 0.838\,{\scriptsize($\pm$.001)} \\
Symbol Tuning~\cite{wei2023symbol} & 0.171 & 0.785 & 0.535 & 0.497\,{\scriptsize($\pm$.042)} & 0.648\,{\scriptsize($\pm$.027)} \\
Flipped-Label Tuning~\cite{agrawal2025spokenlanguageunderstandingunseen} & 0.486 & 0.816 & 0.957 & 0.753\,{\scriptsize($\pm$.034)} & 0.825\,{\scriptsize($\pm$.002)} \\
\methodflip{} (ours)      & 0.594 & 0.854 & 0.962 & 0.803\,{\scriptsize($\pm$.005)} & 0.827\,{\scriptsize($\pm$.004)} \\
\methodsym{}  (ours)   & 0.636 & 0.851 & 0.974 & \textbf{0.820}$^*$\,{\scriptsize($\pm$.005)} & 0.756\,{\scriptsize($\pm$.011)} \\
\midrule
\multicolumn{6}{l}{\textit{SALMONN}} \\
Pre-trained          & 0.048 & 0.049 & 0.027 & 0.041 & 0.134 \\
Instruction Tuning   & 0.085 & 0.074 & 0.678 & 0.279 & 0.684 \\
Symbol Tuning~\cite{wei2023symbol} & 0.031 & 0.070 & 0.142 & 0.081 & 0.168 \\
Flipped-Label Tuning~\cite{agrawal2025spokenlanguageunderstandingunseen} & 0.192 & 0.516 & 0.832 & 0.513 & 0.564 \\
\methodflip{} (ours)  & 0.360 & 0.410 & 0.937 & 0.569 & 0.600 \\
\methodsym{} (ours)   & 0.337 & 0.510 & 0.954 & \textbf{0.600} & 0.601 \\
\bottomrule
\end{tabular}
\end{table}

Table~\ref{tab:trained} shows that under original labels, \methodflip{} preserves canonical alignment competitively with standard Instruction Tuning ($0.817$ vs.\ $0.836$ on AF3; $0.827$ vs.\ $0.838$ on Qwen2-Audio), whereas \methodsym{} elicits a modest drop in performance ($0.762$ on AF3, $0.756$ on Qwen2-Audio, and $0.601$ on \textsc{SALMONN}). This trade reflects the intended design: when the same  tasks are evaluated under symbolic labels, standard Instruction Tuning degrades sharply ($0.508$ on AF3, $0.636$ on Qwen2-Audio, and $0.279$ on \textsc{SALMONN}), whereas \methodsym{} preserves its performance.  

\subsection{Disentangling In-Context Drivers: Definitions versus Demonstrations}
\label{subsec:defvsdemo}

\paragraph{Inference-Time Prompt Variants.}
\begin{figure}
  \centering
  \includegraphics[width=0.85\linewidth]{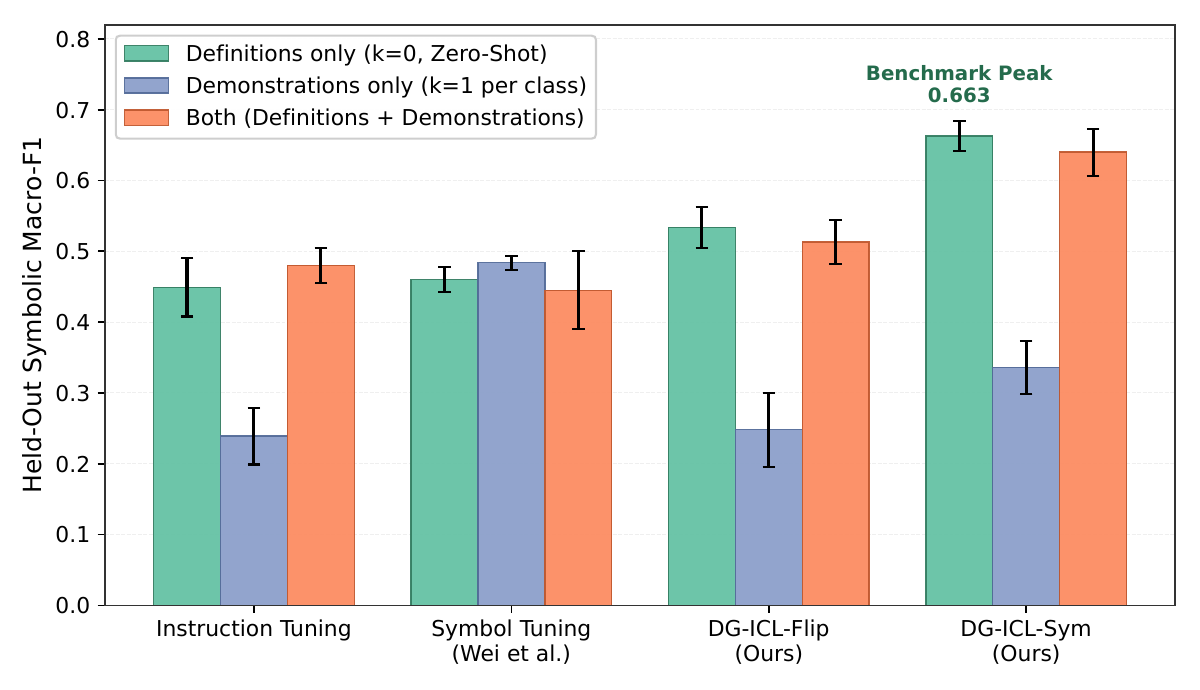}
  \caption{Held-out zero-shot versus few-shot performance (symbolic-label macro-F1 on \textsc{Audio Flamingo~3}) across prompt information regimes. Models are evaluated with natural-language definitions only ($k=0$), one demonstration per class only ($k=1$), or both combined.  Error bars denote standard deviation over three seeds.}
  \label{fig:def-vs-demo}
\end{figure}

To determine which in-context signal drives generalization during evaluation, Figure~\ref{fig:def-vs-demo} evaluates models under three prompt configurations: (1) Definitions only ($k=0$), (2) Demonstrations only ($k=1$ exemplar per class, without definitions), and (3) Both combined.

Two primary conclusions emerge:
\begin{enumerate}
    \item \textbf{Definitions are the primary driver for \method{}:} \methodsym{} achieves peak performance from definitions alone ($0.663$ macro-F1 at $k=0$), outperforming every baseline across all regimes. Appending one demonstration per class provides no additive benefit ($0.640$), whereas removing definitions and providing only demonstrations triggers a sharp drop to $0.336$. \methodflip{} exhibits an identical profile ($0.534$ with definitions vs.\ $0.248$ with demonstrations alone), confirming that dynamic tuning enables the models to rely on definitional semantics rather than exemplars (full per-task results are reported in Table~\ref{tab:def-demo-full}, Appendix~\ref{app:full}).
    
    \item \textbf{Standard baselines cannot substitute exemplars for definitions:} Exemplar-based Symbol Tuning~\cite{wei2023symbol} reaches its best score in its native demonstration regime ($0.484$), but drops to $0.460$ when given definitions alone. Standard Instruction Tuning remains largely flat across regimes ($0.449 \rightarrow 0.480$).  
\end{enumerate}

\paragraph{Training-Time Supervision Variants.}
\begin{table}[t]
\centering
\small
\caption{Instruction-tuning training variants versus \method{} on \textsc{Audio Flamingo~3} (macro-F1 on held-out and trained tasks). Models are varied by the presence of in-context definitions and few-shot demonstrations ($k=1$ per class) during training. Values report mean\,{\scriptsize($\pm$std)} macro F1 over three seeds; best symbolic performance in \textbf{bold}.}
\label{tab:it-ablation}
\begin{tabular}{lcccccc}
\toprule
 & & & \multicolumn{2}{c}{Symbolic labels} & \multicolumn{2}{c}{Original labels} \\
\cmidrule(lr){4-5}\cmidrule(lr){6-7}
Training variant & Def. & Demo. & Held-out & Trained & Held-out & Trained \\
\midrule
Instruction Tuning              & \checkmark & $\times$   & 0.449\,{\scriptsize($\pm$.041)} & 0.508\,{\scriptsize($\pm$.028)} & 0.772\,{\scriptsize($\pm$.002)} & 0.836\,{\scriptsize($\pm$.002)} \\
\quad + demonstrations          & \checkmark & \checkmark & 0.450\,{\scriptsize($\pm$.033)} & 0.445\,{\scriptsize($\pm$.008)} & 0.733\,{\scriptsize($\pm$.003)} & 0.702\,{\scriptsize($\pm$.005)} \\
\quad + demonstrations, no def. & $\times$   & \checkmark & 0.383\,{\scriptsize($\pm$.029)} & 0.519\,{\scriptsize($\pm$.045)} & 0.739\,{\scriptsize($\pm$.005)} & 0.733\,{\scriptsize($\pm$.009)} \\
\midrule
\methodflip{}  (ours)            & \checkmark & $\times$   & 0.534\,{\scriptsize($\pm$.029)} & 0.546\,{\scriptsize($\pm$.053)} & 0.780\,{\scriptsize($\pm$.005)} & 0.817\,{\scriptsize($\pm$.004)} \\
\methodsym{} (ours)                & \checkmark & $\times$   & \textbf{0.663}\,{\scriptsize($\pm$.021)} & \textbf{0.638}\,{\scriptsize($\pm$.044)} & 0.766\,{\scriptsize($\pm$.026)} & 0.762\,{\scriptsize($\pm$.006)} \\
\bottomrule
\end{tabular}
\end{table}

This dominance of definitions extends directly from inference prompts to model training. Table~\ref{tab:it-ablation} investigates the training-time supervisory signals for instruction tuning on \textsc{Audio Flamingo~3}, comparing models trained with definitions only, demonstrations only ($k=1$), or both. Adding demonstrations alongside definitions during training provides no benefit on held-out symbolic labels ($0.450$ vs.\ $0.449$) and degrades canonical-label performance on trained tasks ($0.836 \rightarrow 0.702$) and held-out tasks ($0.772 \rightarrow 0.733$). Omitting definitions during training causes held-out symbolic performance to drop sharply to $0.383$. This confirms our hypothesis for the proposed DG-ICL framework (per-dataset results in Table~\ref{tab:it-ablation-full}, Appendix~\ref{app:full}).

Together, these findings across both inference and training carry substantial practical implications: providing few-shot exemplars across multi-class taxonomies scales prompt length linearly with category count $|L| \times k$ and requires curated examples for every new label. In contrast, natural-language definitions specify category boundaries concisely without requiring labeled exemplars, eliminating the need of in-context examples for SLU.

\subsection{Counterfactual Label Robustness}
\label{subsec:steer}

\begin{figure}
  \centering
  \includegraphics[width=0.85\linewidth]{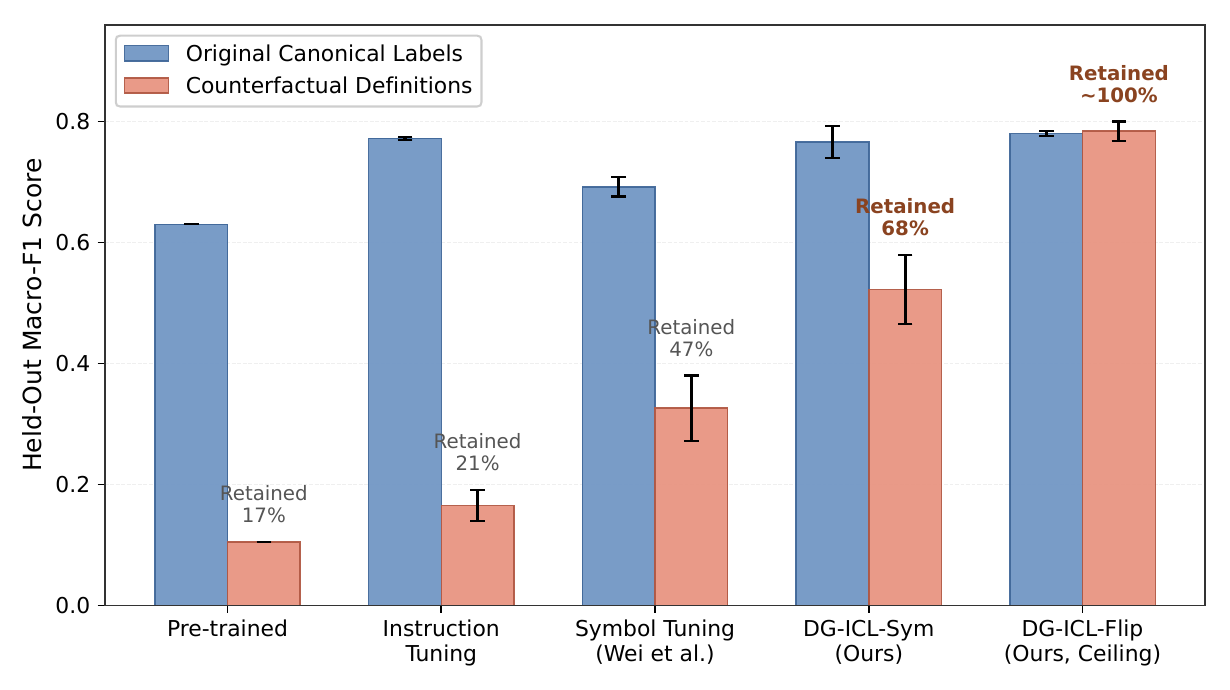}
  \caption{Instruction following ability under counterfactual definition mappings on held-out tasks (macro-F1 on \textsc{Audio Flamingo~3}). Held-out performance is reported before (Original Canonical Labels) and after (Counterfactual Definitions) label-definition modification, alongside the retention percentage labeled above each bar ($\text{Retained} = \text{Counterfactual} / \text{Original}$).  Error bars denote standard deviation over three seeds.}
  \label{fig:steerability}
\end{figure}

A rigorous test of instruction following is whether a model adheres to an in-context definition, where the label semantics \emph{contradict} pre-trained lexical associations. In this counterfactual evaluation, we permute the mapping between canonical label names and definitions on held-out tasks. A model that  interprets definitions should follow the reassignment, whereas one relying on   shortcuts will default to prior lexical meanings. Figure~\ref{fig:steerability} reports held-out macro-F1 before (Original) and after (Counterfactual) permutation, alongside the retained performance ratio ($\text{Counterfactual} / \text{Original}$).

Under canonical labels, standard Instruction Tuning and \methodsym{} achieve comparable performance ($\approx 0.77$ macro-F1), which masks underlying shortcut reliance. Under contradictory definitions, however, models diverge sharply: the pre-trained baseline and Instruction Tuning collapse to \textbf{17\%} ($0.630 \rightarrow 0.105$) and \textbf{21\%} ($0.772 \rightarrow 0.165$) of their canonical performance, respectively. Exemplar-based Symbol Tuning retains \textbf{47\%} ($0.692 \rightarrow 0.326$). In contrast, \methodsym{} retains \textbf{68\%} ($0.766 \rightarrow 0.522$), actively tracking contradictory definitions. Finally, \methodflip{}, optimized directly on label permutations, provides a matched ceiling by retaining \textbf{100\%} ($0.780 \rightarrow 0.784$).

These results confirm that \methodsym{}  encourages in-context definition following capabilities rather than relying on superficial label cues.

\subsection{User-Defined Speech Clustering into Dynamic Categories}
\label{subsec:cluster}

\begin{figure}
  \centering
  \includegraphics[width=0.95\linewidth]{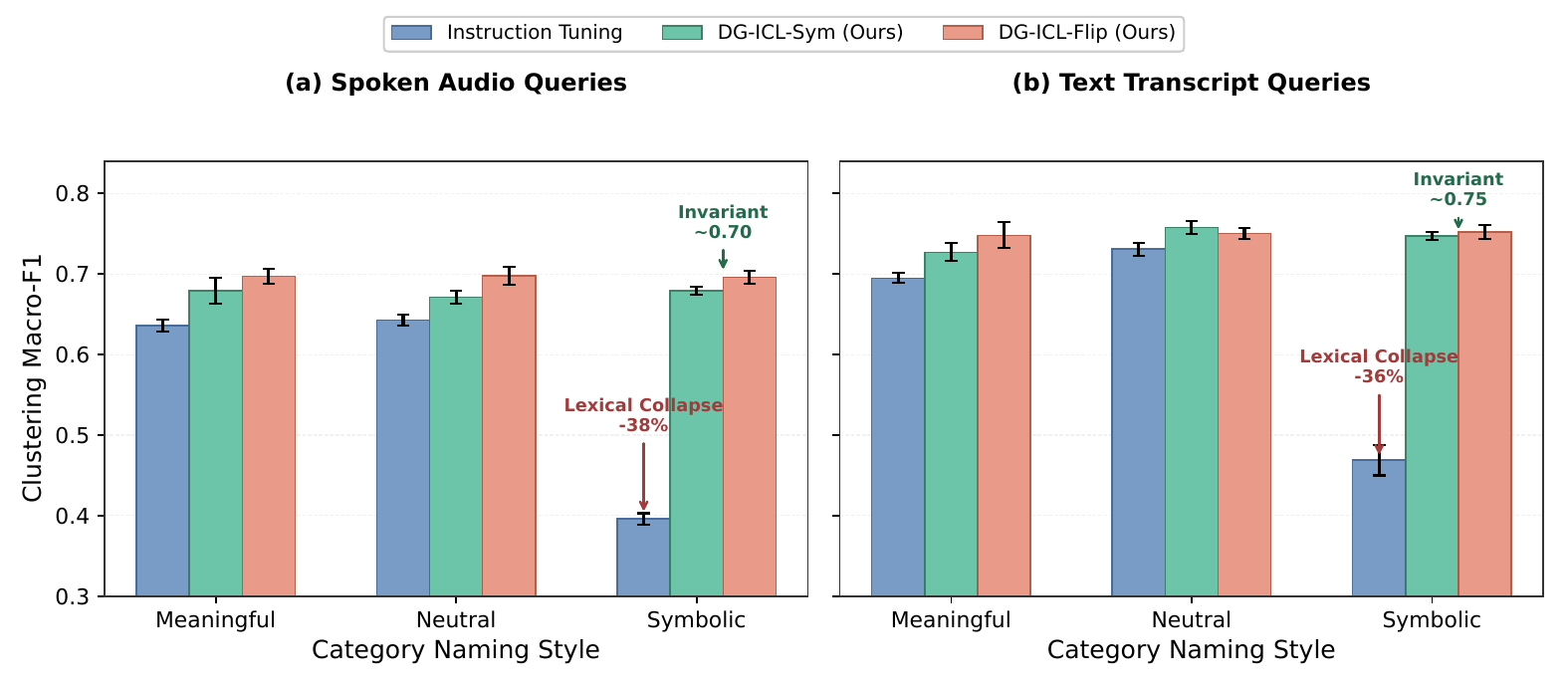}
  \caption{Zero-shot speech clustering performance into user-defined categories across label naming styles (Meaningful, Neutral, Symbolic) on \textsc{Audio Flamingo~3}, averaged over Skit-S2I and MInDS-14 (French and Korean): (a) Spoken audio queries; (b) Text transcript queries.   Error bars denote standard deviation over three seeds.}
  \label{fig:speech-clustering}
\end{figure}

We evaluate our primary application: categorizing spoken queries into novel, user-defined taxonomies which are zero-shot solely provided through natural-language definitions. Figure~\ref{fig:speech-clustering} presents clustering macro-F1 across Skit-S2I and MInDS-14 (French and Korean) under Meaningful, Neutral, and Symbolic naming styles for both spoken audio and text transcript queries (per-dataset breakdowns appear in Table~\ref{tab:cluster-full}, Appendix~\ref{app:clustering}).

When categories carry \emph{meaningful} or \emph{neutral} names, \method{} achieves the highest accuracy across both modalities, outperforming standard Instruction Tuning by $+0.061$ on meaningful audio ($0.697$ vs.\ $0.636$) and by $+0.055$ on neutral placeholders ($0.698$ vs.\ $0.643$). Under arbitrary \emph{symbolic} names, however, performance diverges sharply: while Instruction Tuning experiences a severe $38\%$ drop on audio ($0.636 \rightarrow 0.396$) and $36\%$ on text ($0.731 \rightarrow 0.469$), both \methodsym{} and \methodflip{} maintain invariant accuracy ($\sim 0.68\text{--}0.70$ on audio, $\sim 0.75$ on text).

This collapse demonstrates that standard instruction tuning leans heavily on  lexical shortcuts. In contrast, \method{} consistently follows in-context definitions regardless of label names. This robustness replicates across both spoken audio and text queries, confirming that definition grounding operates as an effective, modality-agnostic mechanism.

\subsection{Analysis and Ablation Studies}
\label{subsec:ablation}

We systematically analyze the components of \method{} on \textsc{Audio Flamingo~3}, reporting held-out symbolic macro-F1 unless mentioned otherwise.

\paragraph{Dynamic versus Static Symbol Assignment.}
\begin{figure}
  \centering
  \includegraphics[width=0.92\linewidth]{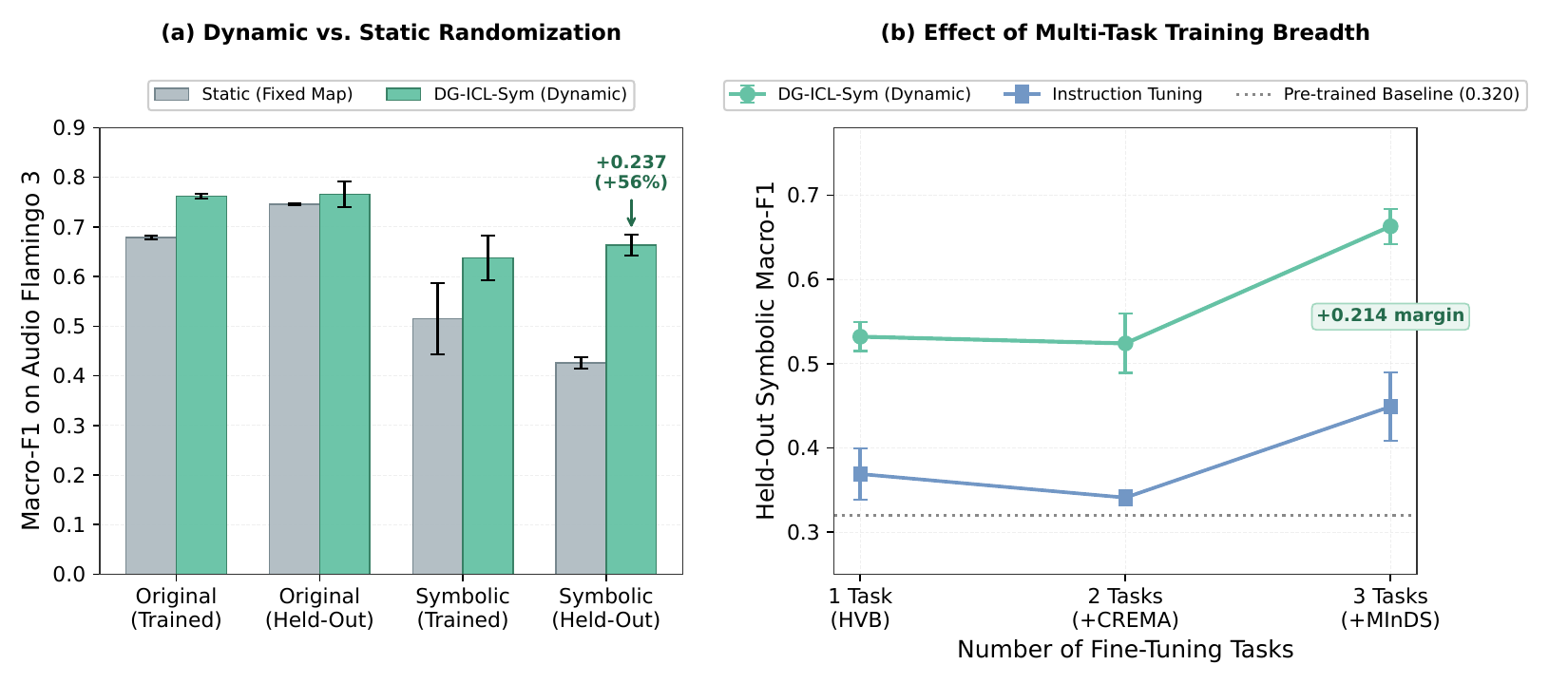}
  \caption{Ablation analysis of core fine-tuning mechanisms on \textsc{Audio Flamingo~3}: (a) Dynamic versus static symbol assignment across original and symbolic label regimes on trained and held-out tasks. Dynamically resampling the label mapping per training instance produces a $+0.237$ absolute gain ($+56\%$) on held-out symbolic evaluation over static substitution ($0.426 \rightarrow 0.663$). (b) Scaling effect of multi-task fine-tuning breadth across one, two, and three training tasks on held-out symbolic macro-F1. Error bars denote standard deviation over seeds.}
  \label{fig:dynamic-breadth}
\end{figure}

Figure~\ref{fig:dynamic-breadth}a compares \methodsym{} against \emph{static symbol tuning}, where the label-to-symbol mapping $\pi : L_{\text{sem}}^T \to L_{\text{sym}}$ is fixed throughout training. Static substitution yields no benefit over standard Instruction Tuning on held-out symbolic labels ($0.426$ vs.\ $0.449$), as the model simply memorizes static associations between acoustic profiles and new tokens without attending to definitions. In contrast, per-instance dynamic resampling $f \sim \mathcal{U}(\mathcal{F}_{\text{sym}})$ disrupts fixed token shortcuts, raising held-out symbolic macro-F1 to \textbf{0.663} ($+0.237$ over static tuning) while maintaining original-label performance ($0.766$ vs.\ $0.746$).

\paragraph{Effect of Multi-Task Training Breadth.}
Figure~\ref{fig:dynamic-breadth}b evaluates how fine-tuning task diversity influences out-of-distribution transfer by varying the number of training tasks from one to three. Across all task counts, \methodsym{} substantially outperforms standard Instruction Tuning ($0.532$, $0.524$, and \textbf{0.663} vs.\ $0.369$, $0.341$, and $0.449$), with the performance margin widening as task diversity increases ($+0.163 \rightarrow +0.183 \rightarrow +0.214$). Multi-task training provides instructional diversity, while \methodsym{} encourages the model to ground predictions in definition text rather than task-specific shortcuts.

\paragraph{Inference-Time Demonstration Scaling (Instruction Tuning and \methodflip{}).}
\begin{figure}
  \centering
  \includegraphics[width=0.92\linewidth]{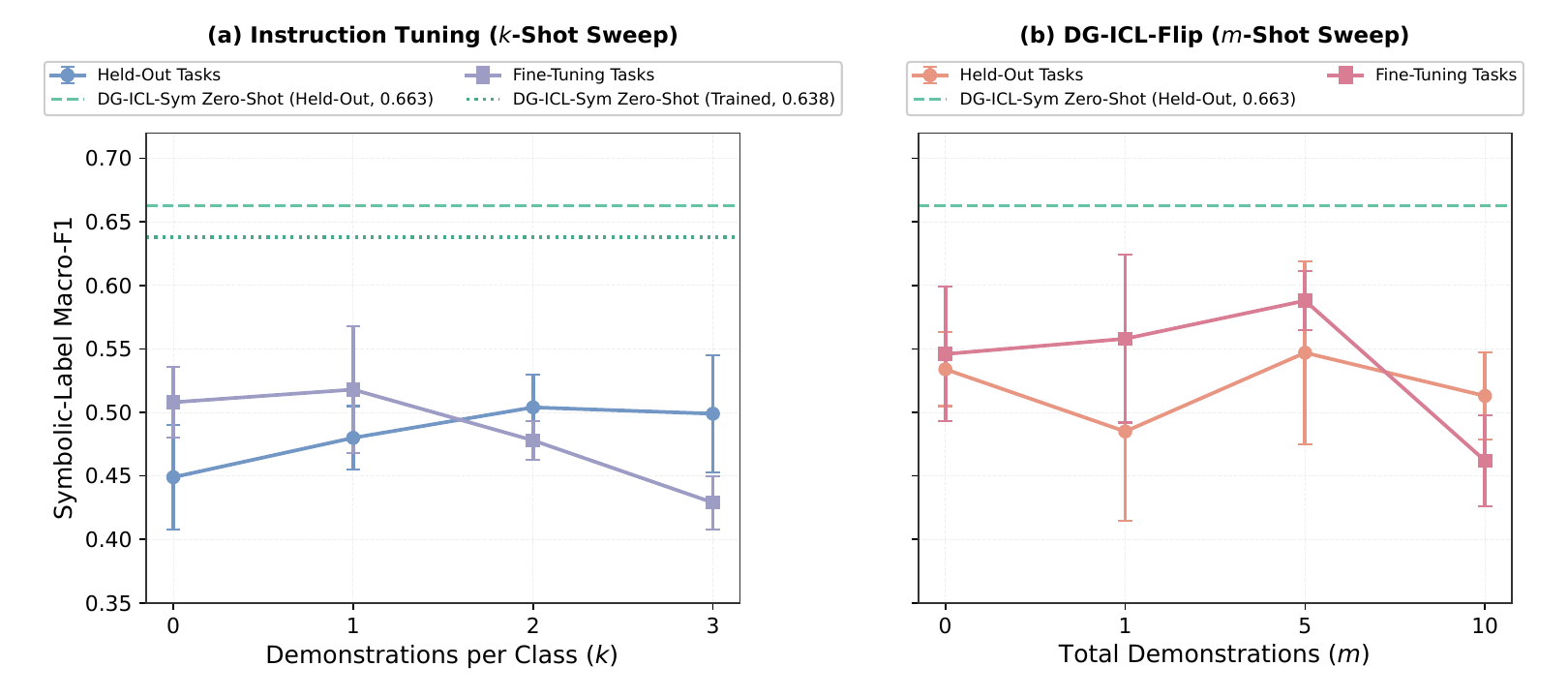}
  \caption{Demonstration scaling sweeps under symbolic-label evaluation on \textsc{Audio Flamingo~3}: (a) Instruction Tuning evaluated with $k \in \{0, 1, 2, 3\}$ demonstrations per class across held-out and fine-tuning tasks, showing a flat plateau that fails to approach \methodsym{} zero-shot performance (dashed horizontal lines); (b) \methodflip{} evaluated with $m \in \{0, 1, 5, 10\}$ total demonstrations across classes. Error bars denote standard deviation over three seeds.}
  \label{fig:demonstration-sweeps}
\end{figure}

We evaluate whether providing few-shot exemplars during inference can compensate for the lack of dynamic training (Figure~\ref{fig:demonstration-sweeps}). For standard Instruction Tuning (Figure~\ref{fig:demonstration-sweeps}a), scaling demonstrations from $k=0$ to $k=3$ per class yields a flat performance on held-out symbolic tasks ($0.449 \rightarrow 0.499$), while performance on trained tasks degrades ($0.508 \rightarrow 0.429$). Similarly, for \methodflip{} (Figure~\ref{fig:demonstration-sweeps}b), sweeping total exemplars $m \in \{1, 5, 10\}$ provides no consistent gain over zero-shot ($0.534$ zero-shot vs.\ $0.485\text{--}0.547$) and 
degrades original-label accuracy ($0.780 \rightarrow 0.75\text{--}0.76$). Crucially, in both cases, demonstration scaling fails to approach \methodsym{} zero-shot performance ($0.663$).

\paragraph{Preservation of Foundational Multimodal Capabilities.}
\begin{table}[t]
\centering
\caption{Preservation of foundational multimodal capabilities post fine tuning. We evaluate out-of-domain spoken question answering (HeySQuAD, token-level F1) and keyword spotting (Google Speech Commands, accuracy on native labels).  Values report mean\,{\scriptsize($\pm$std)} over three seeds.}
\label{tab:capability}
\begin{tabular}{lcc}
\toprule
Method & HeySQuAD (token-F1) & Speech Commands (acc.) \\
\midrule
Pre-trained            & 0.792 & 0.861 \\
Instruction Tuning   & 0.809\,{\scriptsize($\pm$.003)} & 0.946\,{\scriptsize($\pm$.019)} \\
\methodsym{} (ours)     & 0.812\,{\scriptsize($\pm$.006)} & 0.963\,{\scriptsize($\pm$.003)} \\
\methodflip{} (ours) & 0.810\,{\scriptsize($\pm$.001)} & 0.966\,{\scriptsize($\pm$.004)} \\
\bottomrule
\end{tabular}
\end{table}
To understand whether the proposed fine tuning degrades broader capabilities, Table~\ref{tab:capability} benchmarks generative spoken question answering (HeySQuAD, token-level F1)\footnote{Passages are text and questions spoken audio. Exact match and format compliance follow similar trends (\methodsym{} EM $0.694$, compliance $0.857$; Instruction Tuning EM $0.689$, compliance $0.845$).} and acoustic keyword spotting (Google Speech Commands, accuracy on native labels). Both \methodsym{} and \methodflip{} match or  exceed Instruction Tuning (HeySQuAD: $0.812$ and $0.810$ vs.\ $0.809$; Speech Commands: $0.963$ and $0.966$ vs.\ $0.946$), improving over the pre-trained model ($0.792$ and $0.861$). Thus, definition-guided tuning preserves foundational   reasoning while maintaining compatibility with open-ended generative speech understanding.

\paragraph{Sensitivity to Definition Wording.}
\begin{table}[t]
\centering
\caption{Sensitivity to natural-language definition phrasing on held-out symbolic macro-F1. Models are evaluated across four linguistic phrasings of identical category definitions (Original, Paraphrased, Terse, Verbose).    Values report mean\,{\scriptsize($\pm$std)} over three seeds.}
\label{tab:defsens}
\begin{tabular}{lcccc}
\toprule
Method & Original & Paraphrase & Terse & Verbose \\
\midrule
Instruction Tuning & 0.449\,{\scriptsize($\pm$.041)} & 0.416\,{\scriptsize($\pm$.024)} & 0.434\,{\scriptsize($\pm$.033)} & 0.487\,{\scriptsize($\pm$.012)} \\
\methodsym{} (ours)  & 0.663\,{\scriptsize($\pm$.021)} & 0.571\,{\scriptsize($\pm$.042)} & 0.602\,{\scriptsize($\pm$.036)} & 0.606\,{\scriptsize($\pm$.058)} \\
\methodflip{} (ours) & 0.534\,{\scriptsize($\pm$.029)} & 0.552\,{\scriptsize($\pm$.032)} & 0.526\,{\scriptsize($\pm$.037)} & 0.587\,{\scriptsize($\pm$.031)} \\
\bottomrule
\end{tabular}
\end{table}
Table~\ref{tab:defsens} evaluates held-out symbolic macro-F1 across four controlled linguistic phrasings: Original, Paraphrased, Terse, and Verbose. Across all phrasings, \methodsym{} ($0.57\text{--}0.66$) and \methodflip{} ($0.53\text{--}0.59$) consistently outperform Instruction Tuning ($0.42\text{--}0.49$). Moreover, \methodsym{} exhibits mild sensitivity to phrasing ($0.663$ under original definitions vs.\ $0.571\text{--}0.606$ for paraphrased variants).  

\section{Limitations}
\label{sec:limitations}

Our study has several limitations that suggest directions for future research. First, while we validate \method{} across two representative speech-text foundation models (\textsc{Audio Flamingo~3} and \textsc{Qwen2-Audio}) and incorporate \textsc{SALMONN} as a reference. The computational constraints necessitated training \textsc{SALMONN} on a reduced sample budget from a single seed; evaluating on proprietary frontiers or  large scale models remains an important next step. Second, the primary performance advantage of definition-guided tuning emerges when taxonomies are novel, renamed, or symbolic; on original labels and generative QA, \method{} matches rather than exceeds standard instruction tuning, as those regimes can be resolved through familiar lexical priors alone. Third, when evaluating with textual transcripts rather than spoken audio queries, performance on paralinguistic tasks (such as emotion recognition) drops markedly, reflecting the reality that affective information resides in acoustic and prosodic variations that transcription discards. Finally, while \methodflip{} demonstrates that fine-tuning on permuted labels yields robust counterfactual steerability, a formal mechanistic interpretability study of cross-modal attention routing under semantic contradiction remains an open avenue for future investigation.

\section{Conclusion}
\label{sec:conclusion}

Adapting spoken language understanding systems to novel label schemes has traditionally required either data re-annotation or few-shot demonstrations that substantially expand prompt context. In this work, we introduced \methodexpbold{} (\method{}), a fine-tuning framework that enables speech-text models to generalize unseen classification tasks from natural-language definitions alone. By dynamically decoupling labels from fixed lexical names and randomizing definitions per training instance, \method{} mitigates semantic shortcut learning and encourages cross-modal attention to bind acoustic features directly to in-context textual definitions.

Our evaluations across three speech-text backbones (\textsc{Audio Flamingo~3}, \textsc{Qwen2-Audio}, and \textsc{SALMONN}) yield four key findings. First, \methodsym{} achieves consistent zero-shot gains on held-out benchmarks under novel symbols (e.g., $+0.214$ macro-F1 over standard Instruction Tuning on AF3) while remaining competitive on seen training tasks. Second, natural-language definitions, rather than few-shot demonstrations, serve as the primary driver of in-context generalization, with exemplars providing negligible additive utility. Third, counterfactual evaluations confirm instruction steerability: \methodsym{} retains $68\%$ of its canonical performance when definitions contradict pre-trained priors, whereas standard instruction tuning  retains only $21\%$. Fourth, \method{} enables zero-shot speech clustering into user-defined categories with strong robustness across naming styles.

Together, these results indicate that concise textual definitions offer an effective, token-efficient alternative to few-shot demonstrations for controllable speech understanding. By grounding acoustic representations in dynamic in-context definitions rather than static lexical labels, \method{} facilitates speech systems that can be adapted and controlled directly through natural language at inference time.

\clearpage




\bibliographystyle{cas-model2-names}
\bibliography{cas-refs}

\appendix
\section{Dataset Details}
\label{app:datasets}

\begin{table}[t]
\centering
\small
\caption{Complete statistics for all SLU datasets used for fine-tuning and
evaluation. \emph{Cls.} = number of label classes; splits give the number of
utterances. HeySQuAD is generative (no fixed class set); Speech Commands is
evaluation-only (no training split).}
\label{tab:datasets}
\begin{tabular}{llrrrr}
\toprule
Dataset & Task & Cls. & Train & Val & Test \\
\midrule
\multicolumn{6}{l}{\textit{Fine-tuning tasks}} \\
HVB~\cite{wu2021harpervalleybankdomainspecificspokendialog} & Dialogue acts & 18 & 11{,}344 & 1{,}690 & 6{,}121 \\
CREMA-D~\cite{cremad}        & Emotion & 6  & 5{,}967 & 737 & 738 \\
MInDS-14~en~\cite{minds14}   & Intent  & 14 & 1{,}085 & 362 & 362 \\
\midrule
\multicolumn{6}{l}{\textit{Held-out tasks}} \\
Skit-S2I~\cite{skit}        & Intent  & 14 & 9{,}400 & 1{,}045 & 1{,}400 \\
RAVDESS-Song~\cite{ravdess} & Emotion & 6  & 1{,}496 & 264 & 264 \\
MInDS-14~fr~\cite{minds14}  & Intent  & 14 & 327 & 106 & 106 \\
MInDS-14~ko~\cite{minds14}  & Intent  & 14 & 354 & 119 & 119 \\
\midrule
\multicolumn{6}{l}{\textit{Capability probes}} \\
Speech Commands~\cite{speechcommands} & Keyword spotting & 10 & \textendash & 3{,}703 & 4{,}074 \\
HeySQuAD~\cite{heysquad}    & Spoken QA & \textendash & \textendash & 712 & 712 \\
\bottomrule
\end{tabular}
\end{table}

Table~\ref{tab:datasets} gives the full statistics for every dataset. All the audio files are resampled to 16\,kHz, and all splits follow the official partitions.

\paragraph{Fine-tuning tasks.} The training mixture spans three diverse task types: HVB (18 dialogue-act categories from spoken customer-service dialogues), CREMA-D (six acted emotion categories), and MInDS-14~en (14 banking intents), providing multi-task instructional diversity. HVB is multi-label: an utterance may carry multiple dialogue acts, evaluated via averaged binary macro-F1.

\paragraph{Held-out tasks.} Four tasks are strictly held out: Skit-S2I (accented banking intent), RAVDESS-Song (sung emotional speech under heavy prosodic shift), and MInDS-14 French and Korean (cross-lingual intent transfer from English definitions). VoxCeleb and VoxPopuli are excluded because both appear in the pretraining corpora of the underlying foundation models.

\paragraph{Capability probes.} Speech Commands (keyword spotting on native labels) and HeySQuAD (generative spoken question answering scored via token-F1 and exact match) assess whether surrogate fine-tuning preserves general multimodal capabilities outside classification.

\section{Hyperparameters and Implementation}
\label{app:hyperparams}
We fine-tune LoRA adapters with rank $r=8$, scaling $\alpha=32$, and dropout $0.05$ on self-attention projections. For Audio Flamingo~3 and Qwen2-Audio we adapt $q,k,v,o$; for SALMONN we adapt $q,v$ and fine-tune its speech Q-Former and projection. The audio encoders and language model weights remain frozen. We optimize with AdamW at learning rate $1\times10^{-5}$ with a cosine schedule and $100$ warmup steps. Effective batch size is held at $8$ (AF3: per-step batch size 2, accumulation 4; Qwen2-Audio and SALMONN: batch size 1, accumulation 8). AF3 and Qwen2-Audio train for up to $5$ epochs; the 13B SALMONN model is trained for $3$ epochs capped at $2{,}000$ utterances per task ($5{,}085$ total) as a scale ablation. Checkpoints are selected via mean zero-shot macro-F1 on validation splits of fine-tuning tasks. Experiments run on an NVIDIA RTX~A6000 (48\,GB) GPU.

\paragraph{Symbol Vocabulary Construction.}
\label{app:symbols}
To build the symbol vocabulary, we sample random lowercase alphabetic strings and retain those that tokenize to exactly two subword units consistently across prompt positions (standalone, leading space, trailing comma). This yields a pool of well-formed, meaning-neutral symbols of uniform token length, removing tokenization-induced length artifacts across categories.

\section{Prompt Templates}
\label{app:prompts}

All tasks utilize a structured, chat-formatted prompt comprising a \textsc{System} directive and a \textsc{User} query. The \textsc{System} message provides the general task instruction, the category \emph{definition legend} (one bullet per category: \texttt{- <label>: <definition>}), and output formatting guidelines. The \textsc{User} message presents optional few-shot demonstrations ($k > 0$) followed by the acoustic query. The expected target is the category label (or a comma-separated list for multi-label tasks).

\paragraph{Label Modes and Legend Construction.}
The prompt template structure remains identical across all evaluation modes; only the surface token \texttt{<label>} is modified, while the accompanying definitions remain fixed:
\begin{itemize}
    \item \textbf{Original:} Uses canonical semantic names (e.g., \textit{acknowledge}).
    \item \textbf{Symbolic (\methodsym{}):} Replaced by a novel, length-normalized two-token symbol (e.g., \textit{vfeld}).
    \item \textbf{Counterfactual (\methodflip{}):} Replaced by a permuted real label whose conventional meaning contradicts the accompanying definition (e.g., assigning \textit{thanks} to the definition of acknowledgment).
    \item \textbf{No-Legend (Symbol Tuning baseline):} Each bullet is reduced from \texttt{- <label>: <definition>} to simply \texttt{- <label>}, removing all definitional text so category meaning must be inferred strictly from demonstrations.
\end{itemize}

\paragraph{Exemplar and Query Formatting.}
When few-shot demonstrations are enabled ($k > 0$), they are introduced via \texttt{Here are a few examples:} followed by paired inputs and supervisory labels (\texttt{Label: <label>}). The target evaluation instance follows the prompt header, \texttt{"Now analyse this input:"}, accompanied by either the raw speech audio sequence \texttt{<audio>} or text transcript. In zero-shot evaluation ($k = 0$), the demonstration block is omitted entirely.

\paragraph{Full Prompt Walkthrough (HVB Example).}
Below is the annotated prompt template illustrating dialogue-act classification on HarperValleyBank:

\begin{tcolorbox}[colback=gray!4!white, colframe=gray!60!black, arc=1.5mm, top=2pt, bottom=2pt, left=4pt, right=4pt, title=\textbf{Dialogue-Act Classification Prompt (HarperValleyBank)}]
\footnotesize
\textbf{[SYSTEM DIRECTIVE]} \\
You are a dialogue analysis expert for banking conversations. Based on the statement below, identify all applicable dialogue actions from the following options: \\[3pt]
\textbf{[DEFINITION LEGEND]} \\
- \textit{acknowledge}: Shows understanding or receipt of information \\
\quad $\hookrightarrow$ \textit{In \methodsym{}:} \textit{vfeld}: Shows understanding or receipt of information \\
\quad $\hookrightarrow$ \textit{In \methodflip{}:} \textit{answer\_general}: Shows understanding or receipt of information \\
- \textit{answer\_agree}: Expresses agreement \quad \dots\ (15 additional category definitions) \dots \quad - \textit{thanks}: Expressions of gratitude \\[3pt]
\textbf{[OUTPUT GUIDELINES]} \\
Guidelines: Multiple actions can apply to a single statement; list all applicable actions separated by commas. \\[3pt]
\textbf{[USER MESSAGE: Few-Shot Demonstrations ($k > 0$ only)]} \\
Here are a few examples: \\
Text: sure, I can help you with that \quad Label: \textit{acknowledge} \textit{(or \textit{vfeld} / \textit{answer\_general})} \\[3pt]
\textbf{[USER MESSAGE: Target Query ($k = 0$ zero-shot start point)]} \\
Now analyse this input: \textit{<audio>} \quad $\rightarrow$ \quad \textbf{[EXPECTED COMPLETION]} \textit{acknowledge}
\end{tcolorbox}

\section{Full Per-Dataset Results}
\label{app:full}
This appendix reports the per-dataset results underlying the aggregate tables in the main text. Tables~\ref{tab:full-af3} and~\ref{tab:full-qwen} give, for Audio Flamingo~3 and Qwen2-Audio respectively, macro-F1 on every classification task under both original and symbolic labels. Values are mean\,{\scriptsize($\pm$std)} over three seeds; the pre-trained model is a single run. Aggregate seen and held-out scores for the 13B SALMONN scale ablation are provided in Tables~\ref{tab:main} and~\ref{tab:trained} of the main text, with identical cross-task trends.

\begin{table*}[t]
\centering\small
\setlength{\tabcolsep}{4pt}
\caption{Full per-dataset test performance (macro-F1) on \textsc{Audio Flamingo~3} across fine-tuning and held-out evaluation tasks under both original canonical and symbolic label regimes. Flipped-Label denotes Flipped-Label Tuning~\cite{agrawal2025spokenlanguageunderstandingunseen}. Values report mean\,{\scriptsize($\pm$std)} over three random seeds; pre-trained foundation baseline reports a single run.}
\label{tab:full-af3}
\begin{tabular}{llcccccc}
\toprule
Dataset & Mode & Pre-trained & Instr.\ Tuning & Symbol Tuning & Flipped-Label & \methodsym{} & \methodflip{} \\
\midrule
\multicolumn{8}{l}{\textit{Fine-tuning tasks}} \\
\multirow{2}{*}{HVB} & Original & 0.201 & 0.719\,{\scriptsize($\pm$.006)} & 0.288\,{\scriptsize($\pm$.015)} & 0.711\,{\scriptsize($\pm$.006)} & 0.576\,{\scriptsize($\pm$.020)} & 0.690\,{\scriptsize($\pm$.004)} \\
 & Symbolic & 0.083 & 0.204\,{\scriptsize($\pm$.052)} & 0.132\,{\scriptsize($\pm$.015)} & 0.483\,{\scriptsize($\pm$.093)} & 0.491\,{\scriptsize($\pm$.030)} & 0.459\,{\scriptsize($\pm$.076)} \\
\multirow{2}{*}{CREMA-D} & Original & 0.491 & 0.818\,{\scriptsize($\pm$.005)} & 0.347\,{\scriptsize($\pm$.116)} & 0.793\,{\scriptsize($\pm$.010)} & 0.765\,{\scriptsize($\pm$.011)} & 0.782\,{\scriptsize($\pm$.010)} \\
 & Symbolic & 0.504 & 0.729\,{\scriptsize($\pm$.020)} & 0.269\,{\scriptsize($\pm$.035)} & 0.497\,{\scriptsize($\pm$.069)} & 0.604\,{\scriptsize($\pm$.088)} & 0.482\,{\scriptsize($\pm$.123)} \\
\multirow{2}{*}{MInDS-14~en} & Original & 0.859 & 0.970\,{\scriptsize($\pm$.004)} & 0.952\,{\scriptsize($\pm$.004)} & 0.978\,{\scriptsize($\pm$.002)} & 0.946\,{\scriptsize($\pm$.028)} & 0.980\,{\scriptsize($\pm$.002)} \\
 & Symbolic & 0.415 & 0.593\,{\scriptsize($\pm$.048)} & 0.588\,{\scriptsize($\pm$.084)} & 0.622\,{\scriptsize($\pm$.072)} & 0.819\,{\scriptsize($\pm$.094)} & 0.696\,{\scriptsize($\pm$.095)} \\
\midrule
\multicolumn{8}{l}{\textit{Held-out tasks}} \\
\multirow{2}{*}{Skit-S2I} & Original & 0.831 & 0.909\,{\scriptsize($\pm$.004)} & 0.922\,{\scriptsize($\pm$.001)} & 0.909\,{\scriptsize($\pm$.007)} & 0.917\,{\scriptsize($\pm$.003)} & 0.914\,{\scriptsize($\pm$.004)} \\
 & Symbolic & 0.462 & 0.503\,{\scriptsize($\pm$.114)} & 0.630\,{\scriptsize($\pm$.041)} & 0.728\,{\scriptsize($\pm$.059)} & 0.825\,{\scriptsize($\pm$.053)} & 0.661\,{\scriptsize($\pm$.045)} \\
\multirow{2}{*}{RAVDESS-Song} & Original & 0.490 & 0.609\,{\scriptsize($\pm$.011)} & 0.372\,{\scriptsize($\pm$.046)} & 0.552\,{\scriptsize($\pm$.008)} & 0.560\,{\scriptsize($\pm$.023)} & 0.557\,{\scriptsize($\pm$.015)} \\
 & Symbolic & 0.095 & 0.288\,{\scriptsize($\pm$.075)} & 0.123\,{\scriptsize($\pm$.043)} & 0.421\,{\scriptsize($\pm$.066)} & 0.492\,{\scriptsize($\pm$.075)} & 0.469\,{\scriptsize($\pm$.055)} \\
\multirow{2}{*}{MInDS-14~fr} & Original & 0.763 & 0.925\,{\scriptsize($\pm$.008)} & 0.918\,{\scriptsize($\pm$.007)} & 0.936\,{\scriptsize($\pm$.015)} & 0.923\,{\scriptsize($\pm$.039)} & 0.939\,{\scriptsize($\pm$.003)} \\
 & Symbolic & 0.528 & 0.549\,{\scriptsize($\pm$.090)} & 0.668\,{\scriptsize($\pm$.034)} & 0.583\,{\scriptsize($\pm$.011)} & 0.848\,{\scriptsize($\pm$.012)} & 0.572\,{\scriptsize($\pm$.133)} \\
\multirow{2}{*}{MInDS-14~ko} & Original & 0.436 & 0.645\,{\scriptsize($\pm$.011)} & 0.557\,{\scriptsize($\pm$.027)} & 0.708\,{\scriptsize($\pm$.005)} & 0.661\,{\scriptsize($\pm$.041)} & 0.707\,{\scriptsize($\pm$.017)} \\
& Symbolic & 0.193 & 0.457\,{\scriptsize($\pm$.031)} & 0.417\,{\scriptsize($\pm$.061)} & 0.419\,{\scriptsize($\pm$.049)} & 0.488\,{\scriptsize($\pm$.024)} & 0.434\,{\scriptsize($\pm$.057)} \\
\bottomrule
\end{tabular}
\end{table*}
\begin{table*}[t]
\centering\small
\setlength{\tabcolsep}{4pt}
\caption{Full per-dataset test performance (macro-F1) on \textsc{Qwen2-Audio} across fine-tuning and held-out evaluation tasks under both original canonical and symbolic label regimes. Flipped-Label denotes Flipped-Label Tuning~\cite{agrawal2025spokenlanguageunderstandingunseen}. Values report mean\,{\scriptsize($\pm$std)} over three random seeds; pre-trained foundation baseline reports a single run.}
\label{tab:full-qwen}
\begin{tabular}{llcccccc}
\toprule
Dataset & Mode & Pre-trained & Instr.\ Tuning & Symbol Tuning & Flipped-Label & \methodsym & \methodflip \\
\midrule
\multicolumn{8}{l}{\textit{Fine-tuning tasks}} \\
\multirow{2}{*}{HVB} & Original & 0.023 & 0.674\,{\scriptsize($\pm$.004)} & 0.347\,{\scriptsize($\pm$.024)} & 0.649\,{\scriptsize($\pm$.002)} & 0.572\,{\scriptsize($\pm$.018)} & 0.655\,{\scriptsize($\pm$.011)} \\
 & Symbolic & 0.030 & 0.314\,{\scriptsize($\pm$.016)} & 0.171\,{\scriptsize($\pm$.038)} & 0.486\,{\scriptsize($\pm$.115)} & 0.636\,{\scriptsize($\pm$.010)} & 0.594\,{\scriptsize($\pm$.007)} \\
\multirow{2}{*}{CREMA-D} & Original & 0.541 & 0.867\,{\scriptsize($\pm$.005)} & 0.704\,{\scriptsize($\pm$.062)} & 0.856\,{\scriptsize($\pm$.007)} & 0.832\,{\scriptsize($\pm$.008)} & 0.854\,{\scriptsize($\pm$.004)} \\
 & Symbolic & 0.564 & 0.718\,{\scriptsize($\pm$.004)} & 0.785\,{\scriptsize($\pm$.016)} & 0.816\,{\scriptsize($\pm$.036)} & 0.851\,{\scriptsize($\pm$.003)} & 0.854\,{\scriptsize($\pm$.003)} \\
\multirow{2}{*}{MInDS-14~en} & Original & 0.794 & 0.974\,{\scriptsize($\pm$.001)} & 0.893\,{\scriptsize($\pm$.014)} & 0.972\,{\scriptsize($\pm$.010)} & 0.865\,{\scriptsize($\pm$.029)} & 0.973\,{\scriptsize($\pm$.004)} \\
 & Symbolic & 0.511 & 0.877\,{\scriptsize($\pm$.013)} & 0.535\,{\scriptsize($\pm$.113)} & 0.957\,{\scriptsize($\pm$.015)} & 0.974\,{\scriptsize($\pm$.005)} & 0.962\,{\scriptsize($\pm$.008)} \\
\midrule
\multicolumn{8}{l}{\textit{Held-out tasks}} \\
\multirow{2}{*}{Skit-S2I} & Original & 0.795 & 0.851\,{\scriptsize($\pm$.008)} & 0.852\,{\scriptsize($\pm$.009)} & 0.754\,{\scriptsize($\pm$.030)} & 0.798\,{\scriptsize($\pm$.001)} & 0.762\,{\scriptsize($\pm$.026)} \\
 & Symbolic & 0.744 & 0.810\,{\scriptsize($\pm$.015)} & 0.455\,{\scriptsize($\pm$.102)} & 0.733\,{\scriptsize($\pm$.104)} & 0.767\,{\scriptsize($\pm$.064)} & 0.803\,{\scriptsize($\pm$.009)} \\
\multirow{2}{*}{RAVDESS-Song} & Original & 0.620 & 0.660\,{\scriptsize($\pm$.006)} & 0.578\,{\scriptsize($\pm$.011)} & 0.497\,{\scriptsize($\pm$.048)} & 0.520\,{\scriptsize($\pm$.014)} & 0.508\,{\scriptsize($\pm$.045)} \\
 & Symbolic & 0.592 & 0.484\,{\scriptsize($\pm$.028)} & 0.319\,{\scriptsize($\pm$.031)} & 0.597\,{\scriptsize($\pm$.057)} & 0.668\,{\scriptsize($\pm$.034)} & 0.645\,{\scriptsize($\pm$.036)} \\
\multirow{2}{*}{MInDS-14~fr} & Original & 0.611 & 0.848\,{\scriptsize($\pm$.008)} & 0.792\,{\scriptsize($\pm$.034)} & 0.886\,{\scriptsize($\pm$.025)} & 0.765\,{\scriptsize($\pm$.031)} & 0.878\,{\scriptsize($\pm$.021)} \\
 & Symbolic & 0.387 & 0.743\,{\scriptsize($\pm$.036)} & 0.359\,{\scriptsize($\pm$.046)} & 0.850\,{\scriptsize($\pm$.002)} & 0.856\,{\scriptsize($\pm$.025)} & 0.866\,{\scriptsize($\pm$.017)} \\
\multirow{2}{*}{MInDS-14~ko} & Original & 0.186 & 0.531\,{\scriptsize($\pm$.018)} & 0.529\,{\scriptsize($\pm$.018)} & 0.490\,{\scriptsize($\pm$.034)} & 0.451\,{\scriptsize($\pm$.028)} & 0.522\,{\scriptsize($\pm$.009)} \\
 & Symbolic & 0.146 & 0.443\,{\scriptsize($\pm$.035)} & 0.262\,{\scriptsize($\pm$.021)} & 0.447\,{\scriptsize($\pm$.046)} & 0.521\,{\scriptsize($\pm$.017)} & 0.454\,{\scriptsize($\pm$.049)} \\
\bottomrule
\end{tabular}
\end{table*}

Table~\ref{tab:def-demo-full} breaks the definitions-versus-demonstrations comparison (Figure~\ref{fig:def-vs-demo}) down by dataset, for both the fine-tuning and the held-out tasks. Definitions carry sufficient information for both \methodsym{} and \methodflip{}, while adding demonstrations yields no consistent gain across individual tasks.

\begin{table*}[t]
\centering\small
\caption{Per-dataset comparison of in-context information drivers (symbolic-label macro-F1 on \textsc{Audio Flamingo~3}) across fine-tuning and held-out tasks. Models are evaluated with natural-language definitions only ($k=0$), one demonstration per class only ($k=1$), or both combined. Held-out averages correspond to Figure~\ref{fig:def-vs-demo}. Values report mean\,{\scriptsize($\pm$std)} over three random seeds.}
\label{tab:def-demo-full}
\begin{tabular}{llcccc}
\toprule
Dataset & Regime & Instr.\ Tun. & Symbol Tun. & \methodsym{} & \methodflip{} \\
\midrule
\multicolumn{6}{l}{\textit{Fine-tuning tasks}} \\
\multirow{3}{*}{HVB} & Definitions & 0.204\,{\scriptsize($\pm$.052)} & 0.132\,{\scriptsize($\pm$.015)} & 0.491\,{\scriptsize($\pm$.030)} & 0.459\,{\scriptsize($\pm$.076)} \\
 & Demonstrations & 0.132\,{\scriptsize($\pm$.019)} & 0.354\,{\scriptsize($\pm$.054)} & 0.266\,{\scriptsize($\pm$.035)} & 0.149\,{\scriptsize($\pm$.013)} \\
 & Both & 0.250\,{\scriptsize($\pm$.045)} & 0.424\,{\scriptsize($\pm$.018)} & 0.566\,{\scriptsize($\pm$.059)} & 0.394\,{\scriptsize($\pm$.014)} \\
\multirow{3}{*}{CREMA-D} & Definitions & 0.729\,{\scriptsize($\pm$.020)} & 0.269\,{\scriptsize($\pm$.035)} & 0.604\,{\scriptsize($\pm$.088)} & 0.482\,{\scriptsize($\pm$.123)} \\
 & Demonstrations & 0.601\,{\scriptsize($\pm$.031)} & 0.225\,{\scriptsize($\pm$.040)} & 0.106\,{\scriptsize($\pm$.015)} & 0.116\,{\scriptsize($\pm$.024)} \\
 & Both & 0.651\,{\scriptsize($\pm$.053)} & 0.228\,{\scriptsize($\pm$.056)} & 0.600\,{\scriptsize($\pm$.126)} & 0.594\,{\scriptsize($\pm$.140)} \\
\multirow{3}{*}{MInDS-14~en} & Definitions & 0.593\,{\scriptsize($\pm$.048)} & 0.588\,{\scriptsize($\pm$.084)} & 0.819\,{\scriptsize($\pm$.094)} & 0.696\,{\scriptsize($\pm$.095)} \\
 & Demonstrations & 0.211\,{\scriptsize($\pm$.015)} & 0.782\,{\scriptsize($\pm$.003)} & 0.494\,{\scriptsize($\pm$.054)} & 0.212\,{\scriptsize($\pm$.047)} \\
 & Both & 0.652\,{\scriptsize($\pm$.113)} & 0.679\,{\scriptsize($\pm$.100)} & 0.764\,{\scriptsize($\pm$.206)} & 0.619\,{\scriptsize($\pm$.096)} \\
\midrule
\multicolumn{6}{l}{\textit{Held-out tasks}} \\
\multirow{3}{*}{Skit-S2I} & Definitions & 0.503\,{\scriptsize($\pm$.114)} & 0.630\,{\scriptsize($\pm$.041)} & 0.825\,{\scriptsize($\pm$.053)} & 0.661\,{\scriptsize($\pm$.045)} \\
 & Demonstrations & 0.236\,{\scriptsize($\pm$.053)} & 0.519\,{\scriptsize($\pm$.069)} & 0.520\,{\scriptsize($\pm$.098)} & 0.336\,{\scriptsize($\pm$.071)} \\
 & Both & 0.581\,{\scriptsize($\pm$.090)} & 0.584\,{\scriptsize($\pm$.082)} & 0.844\,{\scriptsize($\pm$.088)} & 0.634\,{\scriptsize($\pm$.106)} \\
\multirow{3}{*}{RAVDESS-Song} & Definitions & 0.288\,{\scriptsize($\pm$.075)} & 0.123\,{\scriptsize($\pm$.043)} & 0.492\,{\scriptsize($\pm$.075)} & 0.469\,{\scriptsize($\pm$.055)} \\
 & Demonstrations & 0.288\,{\scriptsize($\pm$.139)} & 0.127\,{\scriptsize($\pm$.082)} & 0.118\,{\scriptsize($\pm$.036)} & 0.189\,{\scriptsize($\pm$.010)} \\
 & Both & 0.234\,{\scriptsize($\pm$.080)} & 0.116\,{\scriptsize($\pm$.071)} & 0.401\,{\scriptsize($\pm$.101)} & 0.212\,{\scriptsize($\pm$.132)} \\
\multirow{3}{*}{MInDS-14~fr} & Definitions & 0.549\,{\scriptsize($\pm$.090)} & 0.668\,{\scriptsize($\pm$.034)} & 0.848\,{\scriptsize($\pm$.012)} & 0.572\,{\scriptsize($\pm$.133)} \\
 & Demonstrations & 0.287\,{\scriptsize($\pm$.056)} & 0.730\,{\scriptsize($\pm$.146)} & 0.419\,{\scriptsize($\pm$.040)} & 0.272\,{\scriptsize($\pm$.055)} \\
 & Both & 0.718\,{\scriptsize($\pm$.039)} & 0.605\,{\scriptsize($\pm$.048)} & 0.753\,{\scriptsize($\pm$.067)} & 0.718\,{\scriptsize($\pm$.033)} \\
\multirow{3}{*}{MInDS-14~ko} & Definitions & 0.457\,{\scriptsize($\pm$.031)} & 0.417\,{\scriptsize($\pm$.061)} & 0.488\,{\scriptsize($\pm$.024)} & 0.434\,{\scriptsize($\pm$.057)} \\
 & Demonstrations & 0.142\,{\scriptsize($\pm$.028)} & 0.560\,{\scriptsize($\pm$.027)} & 0.287\,{\scriptsize($\pm$.083)} & 0.196\,{\scriptsize($\pm$.092)} \\
 & Both & 0.385\,{\scriptsize($\pm$.093)} & 0.475\,{\scriptsize($\pm$.067)} & 0.561\,{\scriptsize($\pm$.050)} & 0.487\,{\scriptsize($\pm$.089)} \\
\bottomrule
\end{tabular}
\end{table*}

Table~\ref{tab:it-ablation-full} reports the per-dataset breakdown for the instruction-tuning supervisory variants (Table~\ref{tab:it-ablation}) across fine-tuning and held-out tasks.

\begin{table*}[t]
\centering\small
\caption{Per-dataset breakdown of instruction-tuning variants (symbolic-label macro-F1 on \textsc{Audio Flamingo~3}) for the ablation study in Table~\ref{tab:it-ablation}. Notations: IT denotes standard Instruction Tuning; IT$+$d denotes training with demonstrations ($k=1$); IT$+$d,nd denotes training with demonstrations without definitions. Values report mean\,{\scriptsize($\pm$std)} over three random seeds.}
\label{tab:it-ablation-full}
\begin{tabular}{lccccc}
\toprule
Dataset & IT & IT$+$d & IT$+$d,nd & \methodflip{} & \methodsym{} \\
\midrule
\multicolumn{6}{l}{\textit{Fine-tuning tasks}} \\
HVB          & 0.204\,{\scriptsize($\pm$.052)} & 0.095\,{\scriptsize($\pm$.044)} & 0.167\,{\scriptsize($\pm$.011)} & 0.459\,{\scriptsize($\pm$.076)} & 0.491\,{\scriptsize($\pm$.030)} \\
CREMA-D      & 0.729\,{\scriptsize($\pm$.020)} & 0.698\,{\scriptsize($\pm$.027)} & 0.751\,{\scriptsize($\pm$.007)} & 0.482\,{\scriptsize($\pm$.123)} & 0.604\,{\scriptsize($\pm$.088)} \\
MInDS-14~en  & 0.593\,{\scriptsize($\pm$.048)} & 0.542\,{\scriptsize($\pm$.033)} & 0.640\,{\scriptsize($\pm$.149)} & 0.696\,{\scriptsize($\pm$.095)} & 0.819\,{\scriptsize($\pm$.094)} \\
\midrule
\multicolumn{6}{l}{\textit{Held-out tasks}} \\
Skit-S2I     & 0.503\,{\scriptsize($\pm$.114)} & 0.647\,{\scriptsize($\pm$.098)} & 0.561\,{\scriptsize($\pm$.087)} & 0.661\,{\scriptsize($\pm$.045)} & 0.825\,{\scriptsize($\pm$.053)} \\
RAVDESS-Song & 0.288\,{\scriptsize($\pm$.075)} & 0.296\,{\scriptsize($\pm$.051)} & 0.305\,{\scriptsize($\pm$.027)} & 0.469\,{\scriptsize($\pm$.055)} & 0.492\,{\scriptsize($\pm$.075)} \\
MInDS-14~fr  & 0.549\,{\scriptsize($\pm$.090)} & 0.569\,{\scriptsize($\pm$.045)} & 0.426\,{\scriptsize($\pm$.114)} & 0.572\,{\scriptsize($\pm$.133)} & 0.848\,{\scriptsize($\pm$.012)} \\
MInDS-14~ko  & 0.457\,{\scriptsize($\pm$.031)} & 0.289\,{\scriptsize($\pm$.012)} & 0.240\,{\scriptsize($\pm$.025)} & 0.434\,{\scriptsize($\pm$.057)} & 0.488\,{\scriptsize($\pm$.024)} \\
\bottomrule
\end{tabular}
\end{table*}

\section{User-Defined Clustering Details}
\label{app:clustering}

Each clustering task groups fine-grained benchmark intents into three user-defined categories zero-shot strictly from natural-language definitions. Definitions remain identical across methods and naming styles; only category surface names vary.

\paragraph{Cluster Definitions and Naming Styles.}
The three target categories are specified by the following definitions, identical across naming styles:
\begin{itemize}
    \item \textbf{$c_1$ (Information):} Looking up or reviewing existing account information (balances, past transactions, card limits, addresses). Surface names: \textit{info\_lookup} (Meaningful), \textit{category\_1} (Neutral), \textit{vfeld} (Symbolic).
    \item \textbf{$c_2$ (Action):} Taking action or modifying account setups (transferring funds, activating cards, changing limits, paying bills). Surface names: \textit{account\_action} (Meaningful), \textit{category\_2} (Neutral), \textit{qomr} (Symbolic).
    \item \textbf{$c_3$ (Problem):} Reporting issues or seeking account protection (system errors, card issues, suspected fraud, account freeze, lost card). Surface names: \textit{problem\_report} (Meaningful), \textit{category\_3} (Neutral), \textit{tirsk} (Symbolic).
\end{itemize}

\paragraph{Intent-to-Category Ground-Truth Maps.}
Fine-grained benchmark intents are deterministically mapped to the three target clusters:
\begin{itemize}
    \item \textbf{MInDS-14 (French and Korean):} \\
    $c_1$: \textit{balance}, \textit{latest\_transactions}, \textit{atm\_limit}, \textit{address}, \textit{abroad}. \\
    $c_2$: \textit{pay\_bill}, \textit{direct\_debit}, \textit{cash\_deposit}, \textit{high\_value\_payment}. \\
    $c_3$: \textit{app\_error}, \textit{card\_issues}, \textit{freeze}, \textit{business\_loan}, \textit{joint\_account}.
    \item \textbf{Skit-S2I:} \\
    $c_1$: \textit{branch\_address}, \textit{past\_transactions}, \textit{outstanding\_balance}, \textit{ifsc\_code}, \textit{balance\_enquiry}, \textit{dispatch\_status}. \\
    $c_2$: \textit{activate\_card}, \textit{generate\_pin}, \textit{change\_limit}, \textit{loan\_query}. \\
    $c_3$: \textit{card\_issue}, \textit{unauthorised\_transaction}, \textit{block}, \textit{lost}.
\end{itemize}

\paragraph{Per-Dataset Clustering Results.}
Table~\ref{tab:cluster-full} details per-dataset clustering accuracy underlying Figure~\ref{fig:speech-clustering} across datasets, query modalities, and naming styles.

\begin{table}[t]
\centering\small
\caption{Per-dataset zero-shot speech clustering accuracy across Skit-S2I, MInDS-14 French, and MInDS-14 Korean across Meaningful, Neutral, and Symbolic naming styles for spoken audio and text transcript queries. Values report mean\,{\scriptsize($\pm$std)} over three random seeds.}
\label{tab:cluster-full}
\begin{tabular}{llcccc}
\toprule
Dataset & Style & \methodsym{} & \methodflip{} & Symbol Tuning & Instruction Tuning \\
\midrule
\multicolumn{6}{l}{\textit{(a) Spoken Audio Queries}} \\
\midrule
\multirow{3}{*}{MInDS-14~fr}
 & Meaningful & 0.569\,{\scriptsize($\pm$.024)} & 0.638\,{\scriptsize($\pm$.018)} & 0.547\,{\scriptsize($\pm$.008)} & 0.550\,{\scriptsize($\pm$.016)} \\
 & Neutral    & 0.572\,{\scriptsize($\pm$.004)} & 0.642\,{\scriptsize($\pm$.027)} & 0.560\,{\scriptsize($\pm$.016)} & 0.560\,{\scriptsize($\pm$.004)} \\
 & Symbolic   & 0.579\,{\scriptsize($\pm$.012)} & 0.629\,{\scriptsize($\pm$.009)} & 0.509\,{\scriptsize($\pm$.028)} & 0.365\,{\scriptsize($\pm$.004)} \\
\midrule
\multirow{3}{*}{MInDS-14~ko}
 & Meaningful & 0.597\,{\scriptsize($\pm$.030)} & 0.602\,{\scriptsize($\pm$.017)} & 0.549\,{\scriptsize($\pm$.017)} & 0.499\,{\scriptsize($\pm$.004)} \\
 & Neutral    & 0.580\,{\scriptsize($\pm$.014)} & 0.630\,{\scriptsize($\pm$.018)} & 0.591\,{\scriptsize($\pm$.017)} & 0.513\,{\scriptsize($\pm$.021)} \\
 & Symbolic   & 0.583\,{\scriptsize($\pm$.010)} & 0.611\,{\scriptsize($\pm$.016)} & 0.602\,{\scriptsize($\pm$.020)} & 0.353\,{\scriptsize($\pm$.018)} \\
\midrule
\multirow{3}{*}{Skit-S2I}
 & Meaningful & 0.870\,{\scriptsize($\pm$.008)} & 0.851\,{\scriptsize($\pm$.016)} & 0.882\,{\scriptsize($\pm$.005)} & 0.858\,{\scriptsize($\pm$.010)} \\
 & Neutral    & 0.861\,{\scriptsize($\pm$.011)} & 0.822\,{\scriptsize($\pm$.007)} & 0.884\,{\scriptsize($\pm$.005)} & 0.858\,{\scriptsize($\pm$.003)} \\
 & Symbolic   & 0.875\,{\scriptsize($\pm$.005)} & 0.849\,{\scriptsize($\pm$.005)} & 0.875\,{\scriptsize($\pm$.003)} & 0.471\,{\scriptsize($\pm$.007)} \\
\midrule
\multicolumn{6}{l}{\textit{(b) Text Transcript Queries}} \\
\midrule
\multirow{3}{*}{MInDS-14~fr}
 & Meaningful & 0.670\,{\scriptsize($\pm$.015)} & 0.704\,{\scriptsize($\pm$.038)} & 0.616\,{\scriptsize($\pm$.024)} & 0.626\,{\scriptsize($\pm$.012)} \\
 & Neutral    & 0.714\,{\scriptsize($\pm$.012)} & 0.708\,{\scriptsize($\pm$.008)} & 0.676\,{\scriptsize($\pm$.012)} & 0.657\,{\scriptsize($\pm$.009)} \\
 & Symbolic   & 0.711\,{\scriptsize($\pm$.022)} & 0.708\,{\scriptsize($\pm$.008)} & 0.676\,{\scriptsize($\pm$.012)} & 0.396\,{\scriptsize($\pm$.008)} \\
\midrule
\multirow{3}{*}{MInDS-14~ko}
 & Meaningful & 0.616\,{\scriptsize($\pm$.024)} & 0.641\,{\scriptsize($\pm$.014)} & 0.622\,{\scriptsize($\pm$.007)} & 0.563\,{\scriptsize($\pm$.014)} \\
 & Neutral    & 0.672\,{\scriptsize($\pm$.007)} & 0.647\,{\scriptsize($\pm$.018)} & 0.658\,{\scriptsize($\pm$.024)} & 0.636\,{\scriptsize($\pm$.022)} \\
 & Symbolic   & 0.636\,{\scriptsize($\pm$.017)} & 0.647\,{\scriptsize($\pm$.021)} & 0.650\,{\scriptsize($\pm$.034)} & 0.468\,{\scriptsize($\pm$.017)} \\
\midrule
\multirow{3}{*}{Skit-S2I}
 & Meaningful & 0.895\,{\scriptsize($\pm$.006)} & 0.900\,{\scriptsize($\pm$.006)} & 0.912\,{\scriptsize($\pm$.007)} & 0.895\,{\scriptsize($\pm$.012)} \\
 & Neutral    & 0.889\,{\scriptsize($\pm$.013)} & 0.894\,{\scriptsize($\pm$.003)} & 0.870\,{\scriptsize($\pm$.006)} & 0.901\,{\scriptsize($\pm$.017)} \\
 & Symbolic   & 0.894\,{\scriptsize($\pm$.012)} & 0.903\,{\scriptsize($\pm$.009)} & 0.873\,{\scriptsize($\pm$.002)} & 0.544\,{\scriptsize($\pm$.035)} \\
\bottomrule
\end{tabular}
\end{table}






\end{document}